%% file: main.tex
\documentclass{article} 

\usepackage[table]{xcolor}

\usepackage{rei_preprint,times}

\input{math_commands.tex}

\usepackage{hyperref}
\usepackage{url}

\usepackage{graphicx}
\usepackage{booktabs}
\usepackage{multirow}
\usepackage{tabularx}
\usepackage{amsmath}
\usepackage{amssymb}
\usepackage{makecell}
\usepackage{wrapfig}
\usepackage{placeins}

\usepackage{amsthm}

\theoremstyle{plain}
\newtheorem{theorem}{Theorem}
\newtheorem{proposition}{Proposition}
\newtheorem{lemma}{Lemma}

\theoremstyle{definition}
\newtheorem{definition}{Definition}

\title{
REI: Learning Ideation from Repository Evolution
}

\author{
  Muyu Zou$^{1,*}$\quad Xiang Feng$^{2,*}$\quad Yufei Xu$^{2}$\quad Shanshan Ye$^{3}$\\[2pt]
  Zulong Chen$^{1}$\quad Lingyu Mu$^{1}$\quad Jinxin Hu$^{1,\dagger}$\quad Jing Zhang$^{2,\dagger}$\\[4pt]
  $^{1}$Alibaba Group\quad $^{2}$School of Computer Science, Wuhan University\\[3pt]
  $^{3}$Mohamed bin Zayed University of Artificial Intelligence\\[3pt]
  {\small $^{*}$Equal contribution.\quad $^{\dagger}$Corresponding authors.}\\[4pt]
  {\footnotesize \texttt{zoumuyu.zmy@alibaba-inc.com}, \texttt{fengxiang\_cs@whu.edu.cn}}\\
  {\footnotesize \texttt{yufei.xu@whu.edu.cn}, \texttt{cassie.ye133@hotmail.com}}\\
  {\footnotesize \texttt{zulong.czl@alibaba-inc.com}, \texttt{moulingyu.mly@alibaba-inc.com}}\\
  {\footnotesize \texttt{jinxin.hjx@alibaba-inc.com}, \texttt{jingzhang.cv@gmail.com}}
}

\begin{document}

\maketitle

\input{sections/abstract}

\input{sections/introduction}
\input{sections/related_work}
\input{sections/method}

\input{sections/experiments}
\input{sections/conclusion}

\input{sections/acknowledgments}
\clearpage
\bibliography{references}
\bibliographystyle{authoryear}

\appendix
\clearpage

\input{sections/appendix_theory}

\FloatBarrier
\input{sections/appendix_setup}
\FloatBarrier
\input{sections/appendix_analysis_protocol}
\FloatBarrier
\input{sections/appendix_baselines}
\FloatBarrier
\input{sections/appendix_sensitivity}
\FloatBarrier
\input{sections/appendix_industrial_slices}
\FloatBarrier
\input{sections/appendix_rustc}
\FloatBarrier
\input{sections/appendix_replay}
\FloatBarrier
\input{sections/appendix_cases}
\FloatBarrier
\input{sections/appendix_three_layer}

\FloatBarrier
\input{sections/appendix_annotation}

\FloatBarrier
\input{sections/appendix_multiseed}
\FloatBarrier
\input{sections/appendix_limitations}

\end{document}

%% file: math_commands.tex
\usepackage{amsmath,amsfonts,bm}

\def\eqref#1{equation~\ref{#1}}
\def\Eqref#1{Equation~\ref{#1}}

\def\1{\bm{1}}

\DeclareMathAlphabet{\mathsfit}{\encodingdefault}{\sfdefault}{m}{sl}
\SetMathAlphabet{\mathsfit}{bold}{\encodingdefault}{\sfdefault}{bx}{n}



%% file: sections/abstract.tex
\begin{abstract}
As AI systems become increasingly capable of executing and evaluating experiments, ideation is emerging as a bottleneck in automated scientific and engineering iteration. In many real-world domains, useful ideas are sparse and their validation is costly, making direct online exploration inefficient. We observe that repository evolution provides a scalable source of reusable technical experience: code changes record concrete technical hypotheses, while subsequent revisions, adoption, and reuse provide signals about their utility. Based on this observation, we propose \textbf{REI}, a Repository Evolution-driven Ideation framework that learns reusable ideation policies from repository evolution, injecting domain-specific technical experience into model parameters and reducing reliance on costly trial-and-error. We further identify a hidden failure mode in quality-driven ideation optimization: \textbf{semantic diversity collapse}. Although generated ideas can remain lexically diverse, they increasingly concentrate in a small number of dominant method families, limiting cross-task generalization and long-tail exploration. To address this problem, we introduce \textbf{S}emantic \textbf{P}rior \textbf{O}ptimization and \textbf{D}ebiasing (\textbf{SPOD}), a diversity-preserving alignment mechanism that makes method identity explicit, corrects category-level reward bias, and limits redundant accumulation within the same method family while retaining strong representatives. Experiments on public benchmarks and a real-world industrial setting demonstrate improvements in ideation quality, method diversity, cross-task generalization, and long-tail exploration. Our implementation is available at \url{https://github.com/Gabriel-reina/spod}.
\end{abstract}


%% file: sections/introduction.tex
\section{Introduction}

Recursive self-improvement (RSI) is becoming increasingly practical for modern AI systems~\citep{Duan2026TheLA,zhang2026darwin,huang2026rzero}. In a typical RSI loop, a system proposes a modification, executes the corresponding experiment, and retains the change if it yields measurable improvements~\citep{Novikov2025AlphaEvolveAC,RomeraParedes2023MathematicalDF,lange2026shinkaevolve}. As implementation, training, and evaluation become increasingly automated, the primary bottleneck is shifting from \emph{execution} to \emph{ideation}: the central challenge is no longer merely whether a system can run experiments, but whether it can consistently identify high-value directions worth exploring. This challenge is particularly pronounced in high-frequency model and system iteration, where existing workflows increasingly rely on general-purpose language models to propose candidate improvements~\citep{Lu2024TheAS,Yamada2025TheAS,schmidgall-etal-2025-agent,Yang2025RDAgentAL,Team2026AgentXTA}. Such models, however, provide only weak domain-specific priors. In domains such as recommender systems, the value of an idea often becomes clear only after implementation, training, and evaluation, making poorly grounded exploration costly~\citep{Kim2026SelfEvolveRecSR,Wang2026SelfEvolvingRS,Wu2026AgenticRecTuneMW}. Moreover, even when useful directions are discovered through extensive online search, the resulting experience is often tied to a particular task and search trajectory rather than distilled into reusable ideation capability that transfers across tasks.

Repository evolution provides a scalable source of the missing technical experience. A repository records a sequence of concrete engineering hypotheses: each committed modification reflects a specific design choice, while its subsequent acceptance, revision, reuse, or downstream effect provides implicit supervision about the usefulness of that direction. More importantly, repository histories naturally span heterogeneous tasks, system states, and evolution trajectories, exposing recurring patterns of what kinds of modifications tend to succeed under different conditions. Such histories therefore contain not only implementation traces, but also reusable priors over technical problem solving. Compared with learning solely through task-local online exploration, repository-scale evolution offers a way to transform accumulated engineering experience into transferable ideation knowledge~\citep{ouyang2026reasoningbank,zhang2026agentic}, providing a substantially stronger starting point for iterative self-improvement.

However, learning from historical success alone does not ensure broad long-term exploration. We find that ideation policies can gradually become concentrated in a small number of dominant method families, improving same-task reward while weakening cross-task transfer and long-tail coverage (see Fig.~\ref{fig:analysis}). This stems from a self-reinforcing dynamic: once a family is favored, optimization repeatedly doubles down on similar variants, progressively crowding out underrepresented yet viable alternatives. Importantly, this collapse is often hidden at the surface level: ideas may remain lexically diverse while becoming increasingly homogeneous in their underlying methodology (see Fig.~\ref{fig:surface_method}). Reinforcement learning can amplify this effect, and stronger verifiers do not fundamentally solve it; they can instead push the policy further toward already favored method families (see Fig.~\ref{fig:verifier_fidelity}). A practical ideation framework therefore needs not only strong technical priors, but also an optimization mechanism that preserves semantic breadth and long-tail exploration throughout learning.

\begin{figure*}[t]
    \centering
    \includegraphics[width=0.65\textwidth]{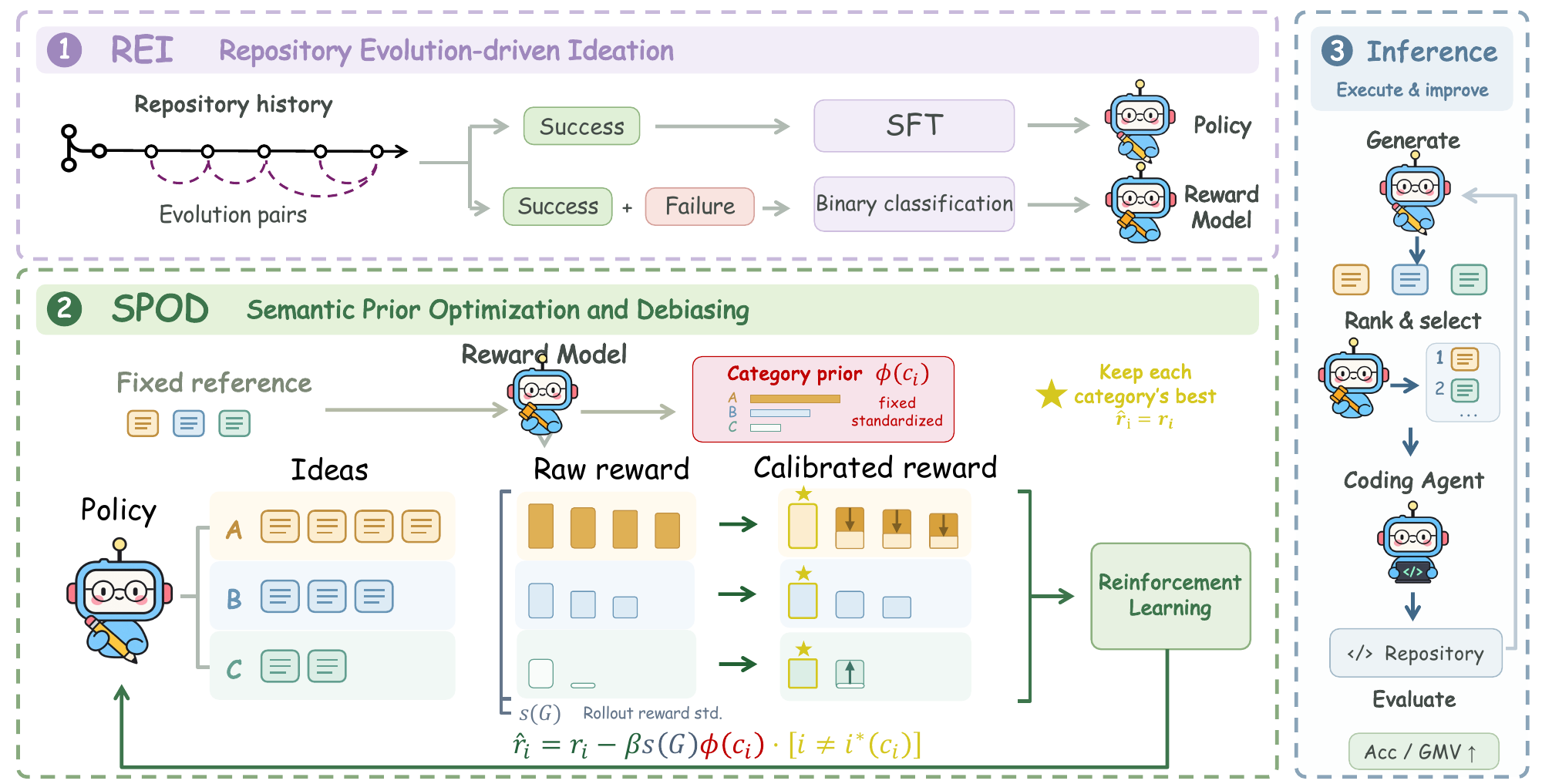}
    \caption{REI learns reusable ideation policies from repository evolution, while SPOD prevents semantic collapse by debiasing method families and retaining strong representatives.}
    \label{fig:overview}
\end{figure*}

To harness repository histories for transferable ideation, we propose \textbf{REI}, a \textbf{R}epository \textbf{E}volution-driven \textbf{I}deation framework that distills historical engineering trajectories into reusable technical priors (Fig.~\ref{fig:overview}). To address the method-level collapse identified above, REI introduces \textbf{SPOD} (\textbf{S}emantic \textbf{P}rior \textbf{O}ptimization and \textbf{D}ebiasing) as its core alignment mechanism. SPOD makes method-level semantic structure explicit, corrects category-level reward imbalance, and employs top-1 retention to suppress redundant variants within a dominant family while preserving its strongest representative, thereby preserving broader exploration without discarding high-quality directions. We evaluate REI on public benchmarks and a real-world industrial recommender system. Across these settings, REI consistently improves ideation quality, semantic method diversity, cross-task generalization, and long-tail exploration. In end-to-end deployment on the industrial recommender system, REI outperforms frontier general-purpose models and human experts, while delivering sustained gains under continual iteration.

Our main contributions are as follows:
\begin{itemize}
    \item \textbf{A repository-evolution framework for ideation.} We propose REI, which learns reusable ideation policies from repository evolution for continual model and system improvement.
    \item \textbf{A diagnosis of hidden method-level collapse.} We reveal a hidden form of method-level collapse in quality-driven ideation optimization: method diversity can collapse even when token entropy and lexical diversity remain relatively stable, and stronger verifiers can further exacerbate this effect.
    \item \textbf{SPOD for diversity-preserving alignment.} We propose SPOD, which uses method-level semantic annotation, category-aware reward correction, and top-1 retention to mitigate method-level collapse during policy optimization.
    \item \textbf{Validation on public and industrial domains.} We evaluate REI and SPOD on public benchmarks and a real-world industrial recommender system, showing consistent gains in ideation quality, diversity, cross-task generalization, and long-tail exploration. In the industrial setting, they further yield sustained gains over continual iteration, including a +4.86\% GMV lift and a +1.82\% advertising revenue lift.
\end{itemize}

%% file: sections/related_work.tex
\section{Related Work}
\subsection{Agentic Automation of Engineering Iteration}

AI is increasingly used to automate engineering iteration, spanning code generation, experiment execution, literature-informed proposal generation, and closed-loop optimization~\citep{Guo2024DSAgentAD,Hong2024DataIA,trirat2025automlagent,nam2025mlestar,hollmann2023large,tang2025airesearcher}. More recent systems further incorporate the resulting interaction trajectories into reinforcement learning, with the goal of turning online trial-and-error into improved ideation capability. In recommender systems, for example, agentic pipelines have been developed to iteratively refine models and strategies through repeated interaction with online feedback~\citep{10.1145/3770854.3785688}. Despite these advances, most existing approaches still rely on a general-purpose LLM as the initial ideation policy, leaving domain-specific technical knowledge to be acquired through costly task-local exploration. REI instead learns this knowledge from repository evolution, turning accumulated engineering trajectories into reusable priors before online iteration begins.

\subsection{Diversity and Exploration in RL Alignment}

A related line of work studies diversity preservation during RL alignment, often building on group-relative policy optimization~\citep{Shao2024DeepSeekMathPT} and introducing exploration controls at the token or policy-update level. Entropy regularization explicitly adds an entropy bonus to discourage premature policy concentration~\citep{Schulman2017ProximalPO}, while entropy-based advantage adjustment assigns stronger learning signals to high-entropy tokens~\citep{Cheng2025ReasoningWE}. Other approaches modify the policy update rule itself. CISPO clips only the importance-sampling weight while retaining the token-level policy gradient~\citep{Chen2025MiniMaxM1ST}; DAPO relaxes the clipping constraint to allow initially low-probability positive tokens to receive larger updates~\citep{yu2025dapo}; and STEER reweights updates according to their predicted effect on entropy, suppressing updates that would induce excessive entropy reduction~\citep{hao-etal-2026-rethinking}. These methods primarily preserve exploration in the token-level policy distribution. In contrast, SPOD focuses on semantic diversity at the method-family level: it explicitly represents method identity and corrects category-level reward imbalance to reduce excessive concentration on a small set of dominant solution families.


%% file: sections/method.tex
\section{Method}
\subsection{Learning Ideation Experience from Repository Histories}

Rather than acquiring ideation ability primarily through costly online trial-and-error, we learn an ideation policy directly from repository evolution. Repository histories provide scalable supervision by recording concrete modifications together with their subsequent empirical outcomes, allowing accumulated engineering experience to be converted into training signals for ideation.


For each target system, we extract training trajectories from its repository history, each comprising a scenario context, a proposed modification, and its observed outcome. We train the ideation policy in two stages. First, we perform supervised fine-tuning on successful historical modifications, grounding the policy in domain-specific technical patterns. Second, we train a verifier on both successful and failed historical outcomes to predict modification utility, and use it to refine the policy via reinforcement learning, optimizing toward directions with stronger empirical gains. During downstream inference, the policy samples candidate ideas that are ranked by the verifier; the top proposals are then dispatched to a coding agent for automated implementation (Fig.~\ref{fig:overview}).

\subsection{Analysis: Diversity Collapse in Idea Generation}

Learning from repository histories provides the ideation policy with a stronger starting point, yet subsequent alignment still determines how exploration is distributed across method families. During iterative optimization, some families are easier to reinforce than others, potentially narrowing the range of solution directions explored by the policy.

This effect is particularly relevant to recommender systems. For example, modifying the loss formulation can quickly improve the current offline metric and is therefore readily rewarded by a verifier, whereas introducing multi-interest user modeling~\citep{Zhou2017DeepIN,Lv2026DAIANDA} may yield smaller initial gains despite its potential value over continued iteration. Repeatedly favoring directions with stronger short-term rewards can therefore concentrate exploration on a narrow subset of method families.

Empirical analysis reveals a systematic collapse in method-level diversity during reinforcement learning. Although the initial policy already prefers certain solution families, standard reinforcement learning amplifies this tendency and concentrates generation on a small number of dominant families. Appendix~\ref{app:analysis_protocol} gives the protocol and metric definitions for the three analyses below.

\begin{wrapfigure}[18]{r}{0.48\textwidth}
  \centering
  \includegraphics[width=\linewidth]{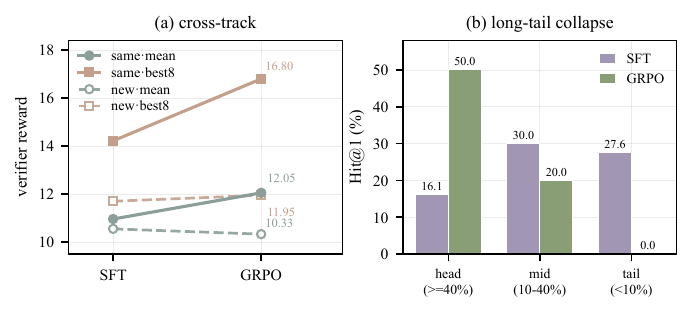}
  \caption{Diversity collapse after RL alignment. \textbf{(a)} Same-task verifier reward improves under GRPO, but cross-task gains disappear. \textbf{(b)} Method families are grouped by their generated share: head ($\ge 40\%$), mid ($10$--$40\%$), and tail ($<10\%$). GRPO improves the head family while degrading the mid and tail families.}
  \label{fig:analysis}
\end{wrapfigure}

\textbf{Long-tail and cross-task performance decline.}
As training progresses, performance on the training task continues to improve. On the same task, GRPO increases both the mean verifier reward and the best-of-8 reward, as shown by the solid lines in Figure~\ref{fig:analysis}(a). However, these gains do not transfer to the out-of-distribution cross-task setting, where the mean reward slightly decreases and the best-of-8 reward changes little (dashed lines in Figure~\ref{fig:analysis}(a)). The discrepancy is more pronounced across method families. When method families are grouped by their share among generated candidates and Hit@1 is measured within each group, GRPO improves the head family while degrading the mid- and tail-frequency families (Figure~\ref{fig:analysis}(b)). Thus, improvements on the training objective can coincide with weaker cross-task generalization and reduced coverage of long-tail method families.

\begin{figure}[!ht]
  \centering
  \vspace{-2mm}
  \includegraphics[width=0.52\textwidth]{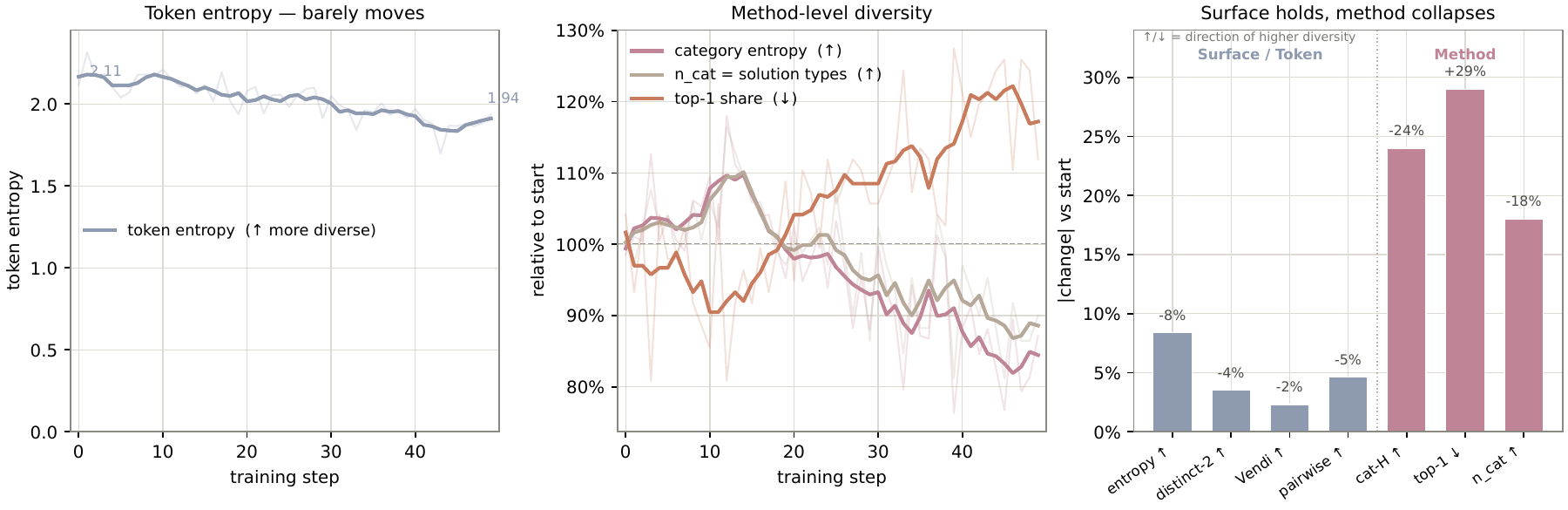}
  \vspace{-3mm}
  \caption{Surface diversity holds while method diversity collapses.
  \emph{Left}: token entropy over training ($2.11 \to 1.94$; axis starts at
  zero). \emph{Middle}: method-level trajectories, normalized to step $0$.
  \emph{Right}: magnitude of change at final checkpoint relative to SFT
  initialization; arrows
  mark higher diversity.}
  \vspace{-2.5mm}
  \label{fig:surface_method}
\end{figure}

\textbf{Lexical diversity does not explain the collapse.}
The degradation in cross-task and long-tail performance suggests that exploration becomes increasingly concentrated during training. However, this concentration cannot be explained by a corresponding collapse in token- or lexical-level diversity. Token entropy exhibits only a marginal decline throughout training (Figure~\ref{fig:surface_method}, left), while lexical dissimilarity changes by less than 5\% relative to the SFT initialization (Figure~\ref{fig:surface_method}, right).

Instead, the collapse is substantially more pronounced at the level of method semantics. An offline LLM maps each generated idea into a method taxonomy $\mathcal{C}$. During training, category entropy and the number of distinct method families $n_{\mathrm{cat}}$ decrease steadily, while the share of the most frequent method family increases monotonically (Figure~\ref{fig:surface_method}, middle). At the final checkpoint, lexical statistics change by only $2$--$8\%$, whereas method-level statistics change by $18$--$29\%$. The policy therefore remains relatively diverse in surface realization while becoming increasingly concentrated in its underlying method families.

\begin{wrapfigure}{r}{0.5\textwidth}
  \centering
  \includegraphics[width=\linewidth]{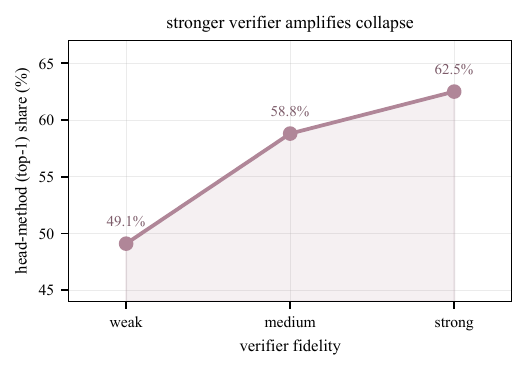}
  \caption{A higher-fidelity verifier accelerates
the collapse: the head-method (top-1) share rises
monotonically across the weak/medium/strong
verifiers (fidelity 0.55/0.67/0.83).}
  \vspace{-5mm}
  \label{fig:verifier_fidelity}
\end{wrapfigure}

\textbf{Stronger optimization further amplifies the collapse.}
Another possible explanation is that the observed concentration is primarily caused by exposure imbalance in the training data or verifier noise. The experimental results do not support this interpretation. Rebalancing the training samples across method categories leaves the collapse largely unchanged. More importantly, improving verifier fidelity, measured as pairwise ranking accuracy between verifier scores and realized gains, does not mitigate the effect; instead, it further increases method-level concentration. As shown in Figure~\ref{fig:verifier_fidelity}, the head-method share rises monotonically from 49.1\% to 58.8\% to 62.5\% as verifier fidelity increases.

Under group-relative updates such as GRPO, category-level reward differences can become self-reinforcing. Candidates from method families that consistently receive higher rewards are more likely to obtain positive advantages, causing subsequent policy updates to further favor these families. Higher-fidelity verification may amplify this effect by preserving the current reward ordering more faithfully. Taken together, these results indicate that the observed collapse occurs primarily at the level of method families rather than surface language. Preserving token-level or lexical diversity is therefore insufficient for maintaining broad exploration. Effective ideation alignment should explicitly account for method-level semantics and reduce systematic reward advantages that repeatedly concentrate optimization on already favored method families.

\subsection{SPOD: Semantic Prior Optimization for Diversity-Preserving Ideation}


To counter method-level concentration, SPOD introduces a three-step alignment pipeline (Fig.~\ref{fig:overview}). For each proposal group, it first assigns method categories through semantic annotation, then computes reward corrections from category reward priors, and finally applies them while exempting each category's highest-reward candidate.


\textbf{Method-Level Semantic Annotation.}
Because the observed collapse occurs at the level of method families rather than surface wording, SPOD explicitly represents method identity. For each candidate idea $y$, we assign a method category $c \in \mathcal{C}$ and prepend the corresponding tag, yielding $\tilde{y} = [\mathrm{TAG}_c] \circ y$, where $\circ$ denotes concatenation. This representation allows subsequent optimization to distinguish candidates by their underlying method family rather than only their lexical form. When a ground-truth taxonomy is available, it is used directly; otherwise, the category is assigned by an offline LLM annotator.
\textbf{Category-Aware Reward Correction.}
Under the same objective, some method categories consistently receive higher rewards than others. Without correction, these category-level reward differences can cause group-relative optimization to repeatedly favor the same method families. SPOD compensates for this effect by adjusting candidate rewards according to the reward prior of their corresponding method category.

A fixed reference set $\mathcal{D}_{\mathrm{ref}}$ of categorized ideas is used to estimate the category-level reward prior. Let $R_i = R(x, y_i)$ denote the raw reward of candidate $y_i$ with category label $c_i$. For a sampled candidate group $G = \{y_1, \dots, y_K\}$ with reward standard deviation $s(G)$, the correction is
\begin{equation}
\mu_c
=
\mathbb{E}_{(x,y)\in\mathcal{D}_{\mathrm{ref}},\,\mathrm{cat}(y)=c}
\left[R(x,y)\right],
\qquad
\varphi(c)
=
\frac{\mu_c-\bar{\mu}}{\sigma_\mu},
\qquad
\Delta_i
=
\beta\,s(G)\,\varphi(c_i),
\end{equation}
where $\bar{\mu}$ and $\sigma_\mu$ denote the mean and standard deviation of the category-wise average rewards, respectively, and $\beta > 0$ controls the correction strength. A larger $\varphi(c)$ corresponds to a stronger positive reward prior for category $c$.

The corrected reward is
\begin{equation}
\tilde{R}_i = R_i - \Delta_i.
\end{equation}


Candidates from categories with stronger positive reward priors receive larger downward corrections, thereby leveling the playing field across method families.

Inspired by debiasing techniques in recommender systems~\citep{Chen2020BiasAD,Gao2023AlleviatingME,Gao2022CIRSBF,Lin2024HowDR}, this mechanism decouples category-level bias from individual candidate quality, thereby neutralizing systematic family advantages while keeping candidate-level quality competition intact. The following theorem formalizes this effect.

\begin{theorem}
In the calibrated reward formulation $r_i' = r_i - \beta \cdot s(G) \cdot \varphi(c_i)$, the normalized fixed-dispersion score $\widetilde{A}_i$ satisfies
\begin{equation}
\widetilde{A}_i
=
\underbrace{
\frac{\delta q_i - \overline{\delta q}_G}{s(G)}
}_{A_i^{\mathrm{intra}}}
+
\left(
1-\frac{\beta \cdot s(G)}{\sigma_\mu}
\right)
\underbrace{
\frac{\mu_{c_i}-\bar{\mu}_G}{s(G)}
}_{A_i^{\mathrm{inter}}},
\end{equation}
where $A_i^{\mathrm{intra}}$ is the category-centered quality term and $A_i^{\mathrm{inter}}$ is the category-mean reward term. The correction leaves the former unchanged and linearly contracts the latter.
\end{theorem}

The theorem shows that category-aware reward correction preserves the category-centered quality term while contracting the category-mean reward term. It therefore reduces the component associated with systematic reward differences across method families without weakening candidate-level quality competition. The full statement and proof are provided in Appendix~\ref{sec:theory}.
\textbf{Top-1 Retention.}
Applying the same correction to every candidate within a method category would also suppress its strongest representative together with weaker redundant variants. SPOD therefore exempts the top candidate within each category and applies the correction only to the remaining candidates:
\begin{equation}
i^\star(c)
=
\arg\max_{j\in\mathcal{G}_c} R(x,y_j),
\qquad
\tilde{R}_i
=
R_i
-
\Delta_i \cdot
\mathbb{I}\!\left(i \neq i^\star(c_i)\right),
\end{equation}
where $\mathcal{G}_c$ denotes the set of candidates in group $G$ that belong to category $c$. The correction is thus concentrated on redundant variants within a method family while preserving its strongest representative, reducing repeated reinforcement of locally similar candidates without removing strong candidates from cross-category competition.


Taken together, SPOD transforms standard reward optimization into a diversity-preserving search: semantic tags expose the underlying method structure, reward correction contracts inter-family bias, and top-1 retention shields the best local candidates from over-penalization. By jointly addressing method-level collapse without discarding high-quality solutions, SPOD maintains a broad and healthy exploration frontier throughout alignment.



%% file: sections/experiments.tex
\begingroup
\section{Experiment}




In this section, we answer three questions: (1) Does REI, coupled with SPOD, deliver superior real-world performance over frontier LLMs and sustain multi-round iteration? (Section~\ref{sec:exp_industrial}, Section~\ref{sec:multi_round}) (2) Can SPOD prevent method collapse and maintain broad exploration across unseen tasks and long-tail domains? (Section~\ref{sec:exp_quality_diversity}, Section~\ref{sec:exp_crosstask}) (3) How does each design component in SPOD affect the final quality-diversity profile? (Section~\ref{sec:exp_ablation})

\subsection{Experimental Setup}

Our experiments combine online and offline evaluations in an industrial AI system, offline evaluation on a public benchmark built from nanoGPT speedrun leaderboard records, and additional offline evaluation on rustc-perf (Appendix~\ref{app:rustc}). Online evaluation uses a production recommender system, where generated ideas undergo the full implementation, training, validation, and deployment pipeline. Offline evaluations use ground-truth method categories for the public benchmarks and LLM-derived reference categories for the industrial system.

\textbf{Evaluation settings.} We use Qoder as the coding agent to implement model-generated ideas. Industrial experiments use Qwen3-8B~\citep{Yang2025Qwen3TR}; public experiments use Qwen2.5-0.5B~\citep{Yang2024Qwen25TR}, with a second public benchmark (rustc-perf, Appendix~\ref{app:rustc}) using Qwen3-1.7B. For each test case, the model generates eight ideas, ranked by the reward model.

\textbf{Metrics and method categories.} For online evaluation, we report Success@$k$, the fraction of prompts with at least one loss-reducing idea among the top-$k$ candidates. For offline evaluation, we assess category alignment (Hit@1, Hit@8, Coverage, TV to Reference Prior) and diversity (Distinct Method Count, Top-Family Share, Max Within-Family Count, Single-Family Collapse Rate). Public categories are ground-truth labels; industrial categories are LLM-extracted from internal repositories. Full metric definitions are in Appendix~\ref{app:setup}.

\textbf{Evaluation slices and baselines.} We report aggregate results, cross-task transfer, and long-tail performance after excluding the dominant category. We compare SPOD with SFT, GRPO~\citep{Shao2024DeepSeekMathPT}, and diversity-oriented baselines, including entropy regularization~\citep{Schulman2017ProximalPO}, entropy-based advantage adjustment~\citep{Cheng2025ReasoningWE}, CISPO~\citep{Chen2025MiniMaxM1ST}, STEER~\citep{hao-etal-2026-rethinking}, and DAPO~\citep{yu2025dapo}. Industrial experiments additionally compare REI with and without SPOD to isolate SPOD's effect under real execution. Baseline details, benchmark construction, implementation, data splits, and extended results appear in Appendix~\ref{app:baselines} and Appendix~\ref{app:setup}.

\subsection{End-to-End Industrial Validation}
\label{sec:exp_industrial}

To assess whether REI improves system performance under actual execution, model-generated proposals are first implemented by a coding agent and then evaluated through training and execution in the industrial environment. As shown in Table~\ref{tab:industrial_validation}, REI achieves substantially higher Success@1 and Success@3 than both human experts and general-purpose LLMs. Specifically, REI-8B reaches 0.68 Success@1 and 0.79 Success@3, compared with 0.32/0.49 for human experts and 0.31/0.46 for GPT-5.5. Incorporating SPOD further improves these metrics to 0.69 and 0.83, respectively. The larger improvement on Success@3 is consistent with the design objective of SPOD, which aims to preserve broader method coverage across multiple candidates while reducing redundant variants within the same method family.

\begin{table*}[t]
\centering
\normalsize

\caption{End-to-end industrial validation. Success@$k$ is judged by true execution after implementation and training. Higher is better for all metrics.}
\label{tab:industrial_validation}
\begin{tabular*}{\textwidth}{@{\extracolsep{\fill}}lcccccc}
\toprule
Metric
& \makecell[c]{Human}
& \makecell[c]{Claude\\4.8-Opus}
& \makecell[c]{DeepSeek\\V4 Pro}
& \makecell[c]{GPT\\5.5}
& \makecell[c]{REI\\8B (base)}
& \makecell[c]{REI\\8B (+SPOD)} \\
\midrule
Success@1 & 0.32 & 0.26 & 0.22 & 0.31 & 0.68 & \textbf{0.69} \\
Success@3 & 0.49 & 0.43 & 0.29 & 0.46 & 0.79 & \textbf{0.83} \\
\bottomrule
\end{tabular*}
\end{table*}

\subsection{Multi-round Iteration}
\label{sec:multi_round}



Real-world optimization typically unfolds over repeated iterations rather than a single round of improvement. To evaluate performance under continual iteration, we conduct experiments in both a live production loop and an offline replay setting based on public speedrun data. In the production loop, starting from the deployed baseline, REI drives 20 consecutive iterations over two weeks and yields a cumulative \textbf{+23.6\%} improvement in Hitrate@200 relative to the initial baseline. Adding SPOD on top of REI brings a further \textbf{+4.2\%} improvement in Hitrate@200. In the final online A/B test, live traffic is randomly split between treatment and control over a fixed observation window. The resulting system further delivers a \textbf{+4.86\%} lift in GMV and a \textbf{+1.82\%} lift in advertising revenue.



For the public speedrun data, an offline replay simulation is conducted by sorting historically accepted improvements within each method family according to their realized gains. At each round, a method family is sampled from the trained model's measured output distribution, and the next unused realized gain from that family is assigned. As shown in Figure~\ref{fig:rsi_main}, GRPO increasingly concentrates on a dominant method family and slows later in the horizon, whereas SPOD maintains broader coverage across method families and continues to accumulate reward. At $T=30$, SPOD achieves a higher cumulative reward than GRPO ($246.5$ vs.\ $197.2$), and the advantage persists under cross-task transfer ($218.1$ vs.\ $171.6$; see Appendix~\ref{app:rsi}). These replay results further indicate that preserving method-family coverage is beneficial for sustained multi-round optimization.


\begin{wrapfigure}[12]{r}{0.5\textwidth}
  \centering
  \vspace{-5mm}
  \includegraphics[width=\linewidth]{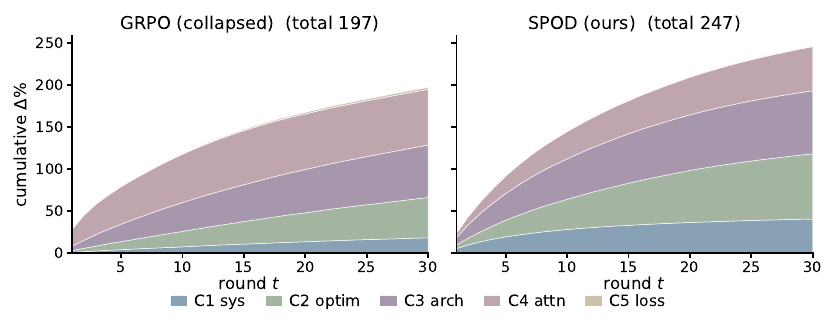}
  \caption{Replay at $T{=}30$: colored bands show cumulative reward by method family (C1--C5; Appendix~\ref{app:setup}). SPOD maintains broader family coverage and achieves higher cumulative reward.}
  \label{fig:rsi_main}
\end{wrapfigure}



\FloatBarrier
\subsection{Overall Quality--Diversity Performance}
\label{sec:exp_quality_diversity}

As shown in Table~\ref{tab:overall_quality}, existing baselines improve only part of the quality profile and do not consistently balance top-ranked accuracy, multi-candidate coverage, and distributional alignment. SPOD achieves the strongest results on all four metrics on the nanoGPT benchmark and shows the same overall trend in the industrial offline evaluation, indicating that SPOD improves category-level candidate quality while bringing the generated method distribution closer to the reference distribution.

The diversity results show a similar pattern. As shown in Table~\ref{tab:overall_diversity}, SPOD achieves a higher Distinct Method Count and lower Top-Family Share and Max Within-Family Count than GRPO and the other post-training baselines, while also maintaining a low Single-Family Collapse Rate. These results indicate that SPOD substantially mitigates semantic concentration toward a small number of dominant method families.




\begin{table*}[!htbp]
\centering
\normalsize
\caption{Overall quality comparison on the industrial offline analysis and the nanoGPT benchmark. Higher is better for Hit@1, Hit@8, and Coverage, while lower is better for TV to Reference Prior.}
\label{tab:overall_quality}
\setlength{\tabcolsep}{5pt}
\begin{tabular*}{\textwidth}{@{\extracolsep{\fill}}lcccccccc}
\toprule
\multirow{2}{*}{Method}
& \multicolumn{4}{c}{Industrial Offline} & \multicolumn{4}{c}{NanoGPT Benchmark} \\
\cmidrule(lr){2-5} \cmidrule(lr){6-9}
& Hit@1$\uparrow$ & Hit@8$\uparrow$ & Coverage$\uparrow$ & TV$\downarrow$
& Hit@1$\uparrow$ & Hit@8$\uparrow$ & Coverage$\uparrow$ & TV$\downarrow$ \\
\midrule
SFT              & 0.687 & 0.960 & 0.960 & 0.238 & 0.391 & 0.783 & 0.707 & 0.297 \\
GRPO             & 0.698 & 0.949 & 0.949 & 0.232 & 0.217 & 0.696 & 0.565 & 0.498 \\
STEER            & 0.730 & 0.944 & 0.944 & 0.189 & 0.217 & 0.739 & 0.641 & 0.475 \\
DAPO             & 0.741 & 0.960 & 0.960 & 0.266 & 0.304 & 0.739 & 0.641 & 0.498 \\
Ent.\ Reg. & 0.698 & 0.860 & 0.860 & 0.283 & 0.217 & 0.783 & 0.652 & 0.524 \\
CISPO            & 0.640 & 0.897 & 0.897 & 0.214 & 0.217 & 0.696 & 0.565 & 0.515 \\
Entro.\ Adv.     & 0.701 & 0.870 & 0.870 & 0.171 & 0.261 & 0.826 & 0.707 & 0.395 \\
SPOD             & \textbf{0.787} & \textbf{0.970} & \textbf{0.970} & \textbf{0.153} & \textbf{0.435} & \textbf{0.913} & \textbf{0.870} & \textbf{0.123} \\
\bottomrule
\end{tabular*}
\end{table*}

\begin{table*}[!htbp]
\centering
\normalsize
\caption{Method diversity on the industrial offline analysis and nanoGPT benchmark. $\uparrow$/$\downarrow$ indicate the preferred direction; bold marks the best value. SPOD is shaded.}
\label{tab:overall_diversity}
\setlength{\tabcolsep}{1.5pt}
\begin{tabular*}{\textwidth}{@{\extracolsep{\fill}}lcccccccc}
\toprule
\multirow{2}{*}{Method}
& \multicolumn{4}{c}{Industrial Offline} & \multicolumn{4}{c}{NanoGPT Benchmark} \\
\cmidrule(lr){2-5} \cmidrule(lr){6-9}
& Distinct$\uparrow$ & Top-Fam$\downarrow$ & Max-Fam$\downarrow$ & Collapse$\downarrow$
& Distinct$\uparrow$ & Top-Fam$\downarrow$ & Max-Fam$\downarrow$ & Collapse$\downarrow$ \\
\midrule
SFT              & 3.054 & 0.420 & 5.247 & \textbf{0.043} & 3.679 & 0.415 & 3.929 & \textbf{0.000} \\
GRPO             & 2.544 & 0.500 & 5.441 & 0.140 & 3.000 & 0.621 & 5.071 & 0.071 \\
STEER            & 2.906 & 0.472 & 4.903 & 0.075 & 3.107 & 0.603 & 4.964 & 0.071 \\
DAPO             & 3.033 & 0.480 & 4.989 & 0.071 & 3.071 & 0.612 & 5.000 & 0.071 \\
Ent.\ Reg. & 3.269 & 0.428 & 4.681 & 0.084 & 3.000 & 0.643 & 5.286 & 0.071 \\
CISPO            & 2.690 & 0.488 & 4.945 & 0.065 & 2.893 & 0.643 & 5.250 & 0.071 \\
Entro.\ Adv.     & 2.961 & 0.476 & 4.659 & 0.090 & 3.429 & 0.522 & 4.393 & 0.036 \\
\rowcolor{gray!12}
SPOD             & \textbf{3.321} & \textbf{0.341} & \textbf{4.021} & 0.056 & \textbf{3.786} & \textbf{0.335} & \textbf{3.571} & \textbf{0.000} \\
\bottomrule
\end{tabular*}
\end{table*}

\subsection{Cross-Task and Long-Tail Evaluation}
\label{sec:exp_crosstask}

\textbf{Cross-task generalization.}
As shown in Table~\ref{tab:cross_longtail_quality} (left), SPOD achieves the highest Hit@8 and Coverage, and the lowest TV to Reference Prior on the nanoGPT cross-task slice. This suggests SPOD better preserves access to relevant method families under task transfer. Although SFT achieves a higher Hit@1, it performs worse on the remaining category-alignment metrics. On the industrial cross-task slice, SPOD achieves the highest Hit@1, Hit@8, and Coverage (Appendix~\ref{app:industrial_slices}, Table~\ref{tab:industrial_cross_longtail}).

\textbf{Long-tail exploration.}
After excluding the dominant head method family on nanoGPT (Table~\ref{tab:cross_longtail_quality}, right), SPOD ties for the highest Hit@1 and achieves the highest Hit@8 and Coverage, with a lower TV than Entro.~Adv. On the industrial long-tail slice, SPOD also achieves the highest Hit@1 and Hit@8, exceeding the strongest baseline for each metric by 7.3 and 6.8 percentage points, respectively. These results show that SPOD's advantages extend to non-head method families.

\begin{table*}[!htbp]
\centering
\normalsize
\caption{Quality comparison on two challenging nanoGPT-benchmark slices: cross-task (left) and long-tail (right). Industrial offline results are reported in Appendix~\ref{app:industrial_slices}.}
\label{tab:cross_longtail_quality}
\setlength{\tabcolsep}{5pt}
\begin{tabular*}{\textwidth}{@{\extracolsep{\fill}}lcccccccc}
\toprule
\multirow{2}{*}{Method}
& \multicolumn{4}{c}{Cross-Task} & \multicolumn{4}{c}{Long-Tail} \\
\cmidrule(lr){2-5} \cmidrule(lr){6-9}
& Hit@1$\uparrow$ & Hit@8$\uparrow$ & Coverage$\uparrow$ & TV$\downarrow$
& Hit@1$\uparrow$ & Hit@8$\uparrow$ & Coverage$\uparrow$ & TV$\downarrow$ \\
\midrule
SFT              & \textbf{0.500} & 0.714 & 0.625 & 0.308 & 0.429 & 0.762 & 0.698 & 0.155 \\
GRPO             & 0.214 & 0.643 & 0.500 & 0.476 & 0.333 & 0.619 & 0.540 & 0.286 \\
STEER            & 0.214 & 0.643 & 0.554 & 0.456 & 0.333 & 0.714 & 0.651 & 0.225 \\
DAPO             & 0.286 & 0.714 & 0.625 & 0.476 & \textbf{0.476} & 0.714 & 0.651 & 0.261 \\
Ent.\ Reg. & 0.357 & 0.643 & 0.536 & 0.449 & 0.333 & 0.667 & 0.587 & 0.207 \\
CISPO            & 0.214 & 0.643 & 0.500 & 0.435 & 0.286 & 0.667 & 0.587 & 0.265 \\
Entro.\ Adv.     & 0.214 & 0.786 & 0.696 & 0.393 & 0.286 & 0.762 & 0.698 & 0.096 \\
SPOD             & 0.286 & \textbf{0.929} & \textbf{0.857} & \textbf{0.134} & \textbf{0.476} & \textbf{0.905} & \textbf{0.857} & \textbf{0.094} \\
\bottomrule
\end{tabular*}
\end{table*}

\subsection{Ablation Study}
\label{sec:exp_ablation}

Table~\ref{tab:ablation_spod} reports a staged ablation of SPOD on the nanoGPT benchmark. The order follows the construction of SPOD: category-aware reward correction requires method-level annotation, while top-1 retention further relies on reward scores within each category. Method-level annotation alone improves Hit@8 and Coverage over GRPO. Adding category-aware reward correction further improves Hit@1, TV to Reference Prior, and the diversity metrics. Finally, incorporating top-1 retention yields the strongest overall quality--diversity performance. The progressive improvements demonstrate the contribution of each component to the complete SPOD formulation.


\begin{table*}[!htbp]
\centering
\normalsize
\caption{Component ablation of SPOD on the nanoGPT benchmark. Ann. denotes method-level annotation, RC denotes category-aware reward correction, and Top1 denotes top-1 retention.}
\label{tab:ablation_spod}
\setlength{\tabcolsep}{4pt}
\begin{tabular*}{\textwidth}{@{\extracolsep{\fill}}lcccccccc}
\toprule
\multirow{2}{*}{Method}
& \multicolumn{4}{c}{Quality}
& \multicolumn{4}{c}{Diversity} \\
\cmidrule(lr){2-5} \cmidrule(lr){6-9}
& Hit@1$\uparrow$ & Hit@8$\uparrow$ & Cov.$\uparrow$ & TV$\downarrow$
& Dist.$\uparrow$ & TopFam$\downarrow$ & MaxInFam$\downarrow$ & Coll.$\downarrow$ \\
\midrule
GRPO               & 0.217 & 0.696 & 0.565 & 0.498 & 3.000 & 0.621 & 5.071 & 0.071 \\
+ Ann.             & 0.217 & 0.826 & 0.707 & 0.431 & 3.000 & 0.643 & 4.607 & 0.036 \\
+ Ann. + RC        & 0.391 & 0.826 & 0.761 & 0.199 & 3.750 & 0.388 & 3.857 & \textbf{0.000} \\
+ Ann. + RC + Top1 & \textbf{0.435} & \textbf{0.913} & \textbf{0.870} & \textbf{0.123}
                   & \textbf{3.786} & \textbf{0.335} & \textbf{3.571} & \textbf{0.000} \\
\bottomrule
\end{tabular*}
\end{table*}



\subsection{Further Analysis}

The appendix reports additional analyses: sensitivity to the debiasing strength $\beta$ and exemption depth $k$ (Appendix~\ref{app:sensitivity}), the cross-task and long-tail slices (Appendix~\ref{app:industrial_slices}), replay simulations and case studies (Appendix~\ref{app:rsi}, \ref{app:case}), diversity dynamics across the token, surface, and method layers (Appendix~\ref{app:three_layer}), consistency across LLM-based extractors under different annotation schemes (Appendix~\ref{app:annot}), multi-seed statistical tests (Appendix~\ref{app:multiseed}) and a second public benchmark (Appendix~\ref{app:rustc}). The main conclusions remain unchanged across these analyses.

\endgroup

%% file: sections/conclusion.tex
\section{Conclusion}

We presented REI, a framework for learning reusable ideation policies from repository histories. We showed that standard reward optimization can cause hidden method-level collapse, even when surface diversity remains largely intact. We therefore introduced SPOD to preserve method coverage during alignment. Experiments on public benchmarks and a real-world industrial recommender system show consistent gains in quality, diversity, and practical effectiveness.

%% file: sections/acknowledgments.tex
\section*{Acknowledgments}

This work was supported in part by the New Generation Artificial
Intelligence-National Science and Technology Major Project
(No.~2025ZD0123602).

%% file: sections/appendix_theory.tex
\section{Theoretical Analysis}
\label{sec:theory}

In this section, we formulate a rigorous theoretical foundation for our proposed calibration framework under Group Relative Policy Optimization (GRPO). We first formalize the problem by characterizing how category-level systemic reward bias propagates through the empirical group-normalization mechanism of GRPO (Section~\ref{subsec:formulation}), establishing an exact dual-channel algebraic decomposition of the policy advantage (Section~\ref{subsec:decomposition}). Subsequently, we prove the structural necessity of coupling the penalty term with the group sample dispersion and a standardized category deviation metric (Section~\ref{subsec:inevitability}). Building upon these results, we establish the uniform linear contraction theorem of the category-mean reward channel and provide formal arguments justifying why a soft shrinkage operator is preferable to hard zero-cancellation (Section~\ref{subsec:shrinkage}). 

\subsection{Problem Formulation and Additive Bias Model}
\label{subsec:formulation}

Let $\mathcal{X}$ denote the prompt domain, and let $\mathcal{Y}$ represent the sequence space of generated responses. Let $\pi_\theta(y \mid x)$ denote the parameterized autoregressive policy. We consider a discrete semantic category universe $\mathcal{C} = \{1, 2, \dots, C\}$ consisting of $C \in \mathbb{N}^+$ distinct categories. We assume the existence of a deterministic semantic classifier $c: \mathcal{Y} \to \mathcal{C}$ that maps each response $y$ to its underlying category label $c(y) \in \mathcal{C}$.

\begin{definition}[Additive Category-Level Bias Model]
\label{def:additive_bias}
For any response $y \in \mathcal{Y}$ belonging to semantic category $c(y) \in \mathcal{C}$, the scalar output $r(y) \in \mathbb{R}$ emitted by the proxy reward model admits an additive decomposition:
\begin{equation}
r(y) = q(y) + b_{c(y)},
\label{eq:additive_bias}
\end{equation}
where $q(y) \in \mathbb{R}$ denotes the latent, instance-level intrinsic quality of response $y$ dictated by its structural correctness and reasoning depth, and $b_{c(y)} \in \mathbb{R}$ represents a fixed-effect systemic bias specific to category $c(y)$, invariant to the microscopic token sequences of $y$.
\end{definition}

To benchmark category-level behavior across the semantic space, we introduce a balanced reference dataset $\mathcal{D}_{\mathrm{ref}}$. For each category $k \in \mathcal{C}$, the category-conditional expected reward is defined as $\mu_k \triangleq \mathbb{E}_{y \sim \mathcal{D}_{\mathrm{ref}}}[r(y) \mid c(y) = k]$. By linearity of the conditional expectation applied to Definition~\ref{def:additive_bias}, we have the exact identity:
\begin{equation}
\mu_k = \bar{q}_k + b_k,
\end{equation}
where $\bar{q}_k \triangleq \mathbb{E}_{y \sim \mathcal{D}_{\mathrm{ref}}}[q(y) \mid c(y) = k]$ characterizes the baseline quality of category $k$ on the reference set. Across the discrete category universe $\mathcal{C}$, the macro-average expectation $\bar{\mu}$ and the inter-category standard deviation $\sigma_\mu$ are defined by:
\begin{equation}
\bar{\mu} \triangleq \frac{1}{C} \sum_{k=1}^C \mu_k, \qquad \sigma_\mu \triangleq \sqrt{\frac{1}{C} \sum_{k=1}^C (\mu_k - \bar{\mu})^2}.
\end{equation}
This induces the dimensionless, standardized category deviation score for category $k$:
\begin{equation}
\varphi(k) \triangleq \frac{\mu_k - \bar{\mu}}{\sigma_\mu}.
\label{eq:phi_def}
\end{equation}
By algebraic construction, $\varphi(k)$ possesses zero sample mean and unit sample variance over $\mathcal{C}$, serving as a scale-invariant metric of category preference.

In standard GRPO training, given a prompt $x \sim \mathcal{X}$, the policy $\pi_\theta$ generates an ensemble of $K$ responses denoted by the ordered tuple $G = (y_1, y_2, \dots, y_K)$. For each sample index $i \in \{1, \dots, K\}$, we abbreviate the instance scalar reward as $r_i \triangleq r(y_i)$, its intrinsic quality as $q_i \triangleq q(y_i)$, and its semantic label as $c_i \triangleq c(y_i) \in \mathcal{C}$. The group sample mean $\bar{r}_G$ and sample standard deviation $s(G)$ are evaluated as:
\begin{equation}
\bar{r}_G \triangleq \frac{1}{K} \sum_{j=1}^K r_j, \qquad s(G) \triangleq \sqrt{\frac{1}{K} \sum_{j=1}^K (r_j - \bar{r}_G)^2}.
\end{equation}
The standardized advantage assigned to response $y_i$ by standard GRPO is defined as:
\begin{equation}
A_i \triangleq \frac{r_i - \bar{r}_G}{s(G)}.
\end{equation}

\subsection{Exact Dual-Channel Advantage Decomposition}
\label{subsec:decomposition}

To characterize the precise influence of category bias in non-asymptotic settings where $K$ is typically small (such as $K=4$ or $K=8$), we bypass any heuristic statistical approximations and establish an exact algebraic decomposition.

For any candidate response $y_i \in G$, we define its individual quality deviation from its category expectation as $\delta q_i \triangleq q_i - \bar{q}_{c_i}$. Across the sampled group $G$, we define the empirical average of individual quality deviations as $\overline{\delta q}_G \triangleq \frac{1}{K} \sum_{j=1}^K \delta q_j$, and the empirical average of sampled category expectations as $\bar{\mu}_G \triangleq \frac{1}{K} \sum_{j=1}^K \mu_{c_j}$.

\begin{proposition}[Dual-Channel Advantage Identity]
\label{prop:advantage_decomp}
For an arbitrary prompt $x$ and any sampled group $G = (y_1, \dots, y_K)$ of size $K \ge 2$, the standard GRPO advantage $A_i$ assigned to sample $y_i$ decomposes identically as:
\begin{equation}
A_i = A_i^{\mathrm{intra}} + A_i^{\mathrm{inter}},
\label{eq:exact_decomp}
\end{equation}
where the intra-category quality competition channel and inter-category category-mean reward channel are defined respectively as:
\begin{equation}
A_i^{\mathrm{intra}} \triangleq \frac{\delta q_i - \overline{\delta q}_G}{s(G)}, \qquad A_i^{\mathrm{inter}} \triangleq \frac{\mu_{c_i} - \bar{\mu}_G}{s(G)}.
\label{eq:channels_def}
\end{equation}
\end{proposition}

\begin{proof}
From the identity $\mu_{c_i} = \bar{q}_{c_i} + b_{c_i}$, we express the category bias directly as $b_{c_i} = \mu_{c_i} - \bar{q}_{c_i}$. Substituting this into Definition~\ref{def:additive_bias} for sample $y_i$ yields:
\begin{equation}
r_i = q_i + b_{c_i} = q_i + (\mu_{c_i} - \bar{q}_{c_i}) = (q_i - \bar{q}_{c_i}) + \mu_{c_i}.
\end{equation}
Recalling the definition $\delta q_i = q_i - \bar{q}_{c_i}$, each observed scalar reward is equivalently expressed as:
\begin{equation}
r_i = \delta q_i + \mu_{c_i}.
\label{eq:r_i_expanded}
\end{equation}
Averaging \Eqref{eq:r_i_expanded} across all $K$ instances within group $G$ and leveraging the finite linearity of summations:
\begin{equation}
\bar{r}_G = \frac{1}{K} \sum_{j=1}^K r_j = \frac{1}{K} \sum_{j=1}^K (\delta q_j + \mu_{c_j}) = \left( \frac{1}{K} \sum_{j=1}^K \delta q_j \right) + \left( \frac{1}{K} \sum_{j=1}^K \mu_{c_j} \right) = \overline{\delta q}_G + \bar{\mu}_G.
\label{eq:r_bar_expanded}
\end{equation}
Subtracting \Eqref{eq:r_bar_expanded} from \Eqref{eq:r_i_expanded} yields the centered reward deviation:
\begin{equation}
r_i - \bar{r}_G = (\delta q_i + \mu_{c_i}) - (\overline{\delta q}_G + \bar{\mu}_G) = (\delta q_i - \overline{\delta q}_G) + (\mu_{c_i} - \bar{\mu}_G).
\end{equation}
Dividing both sides by the group sample dispersion $s(G)$ gives:
\begin{equation}
\frac{r_i - \bar{r}_G}{s(G)} = \frac{\delta q_i - \overline{\delta q}_G}{s(G)} + \frac{\mu_{c_i} - \bar{\mu}_G}{s(G)}.
\end{equation}
Recognizing the left-hand side as $A_i$ and the right-hand terms as $A_i^{\mathrm{intra}}$ and $A_i^{\mathrm{inter}}$ completes the proof.
\end{proof}

Proposition~\ref{prop:advantage_decomp} provides foundational insight into the pathological vulnerability of GRPO. The advantage function consists of two distinct components. The intra-category channel $A_i^{\mathrm{intra}}$ captures genuine capability, quantifying whether instance $y_i$ outperforms its category peers in terms of micro-level generation. Conversely, the inter-category channel $A_i^{\mathrm{inter}}$ serves as a systemic baseline offset. When a category suffers from inflated bias $b_{c_i}$, its reference mean $\mu_{c_i}$ dominates the group baseline $\bar{\mu}_G$. As a result, sub-optimal responses from positively biased categories receive spurious positive advantage scores solely by virtue of their category affiliation, creating an optimization loop that triggers mode collapse.

\subsection{Structural Scale Compatibility of the Penalty Term}
\label{subsec:inevitability}

To suppress the unearned systemic channel $A_i^{\mathrm{inter}}$, one must formulate a calibrated reward $r_i' = r_i - \Delta_i$. We consider a generalized penalty parameterized by $\Delta_i \triangleq f(G) \cdot \phi(c_i)$, where $f(G)$ denotes a group-dependent scale factor and $\phi(c_i)$ represents a category metric.

For the purpose of theoretical analysis, we define the fixed-dispersion normalized score:
\begin{equation}
\widetilde{A}_i \triangleq \frac{r_i' - \bar{r}_G'}{s(G)},
\end{equation}
which isolates the first-order effect of reward calibration on the group-relative numerator while holding the empirical dispersion $s(G)$ fixed. In actual deployment, the calibrated GRPO advantage evaluates as $A_i' \triangleq (r_i' - \bar{r}_G') / s(G')$, where $s(G') \triangleq \sqrt{\frac{1}{K}\sum_j (r_j' - \bar{r}_G')^2}$. When the variance shift induced by calibration is well-behaved, $\widetilde{A}_i$ serves as a direct first-order surrogate for $A_i'$.

\begin{proposition}[Scale Compatibility and Non-Singularity under Collapse]
\label{prop:singularity}
Let $\delta A_i \triangleq \widetilde{A}_i - A_i$ denote the net advantage displacement under the fixed-dispersion score. In order for the calibration mechanism to satisfy:
\begin{enumerate}
    \item Scale compatibility, such that $\delta A_i$ is strictly dimensionless and invariant to linear rescalings of the underlying reward metric;
    \item Non-singularity under mode collapse, such that in the event of severe intra-group homogenization where sample variance vanishes ($s(G) \to 0$), the advantage perturbation remains uniformly bounded ($\lim_{s(G) \to 0} |\delta A_i| < \infty$);
\end{enumerate}
setting $f(G) = \beta \cdot s(G)$ for some fixed scalar $\beta > 0$ and choosing $\phi(c)$ as the standardized deviation $\varphi(c)$ represents a natural and robust formulation.
\end{proposition}

\begin{proof}
Defining the empirical mean of the penalty terms over group $G$ as $\bar{\Delta}_G \triangleq \frac{1}{K}\sum_{j=1}^K \Delta_j$, the calibrated group mean reward is:
\begin{equation}
\bar{r}_G' = \frac{1}{K}\sum_{j=1}^K (r_j - \Delta_j) = \bar{r}_G - \bar{\Delta}_G.
\end{equation}
The fixed-dispersion normalized score takes the form:
\begin{equation}
\widetilde{A}_i = \frac{(r_i - \Delta_i) - (\bar{r}_G - \bar{\Delta}_G)}{s(G)} = \frac{r_i - \bar{r}_G}{s(G)} - \frac{\Delta_i - \bar{\Delta}_G}{s(G)}.
\end{equation}
Subtracting $A_i = (r_i - \bar{r}_G) / s(G)$, the exact advantage displacement simplifies to:
\begin{equation}
\delta A_i = -\frac{\Delta_i - \bar{\Delta}_G}{s(G)}.
\label{eq:delta_A_exact}
\end{equation}
Substituting $\Delta_i = f(G) \cdot \phi(c_i)$ and defining $\bar{\phi}_G \triangleq \frac{1}{K}\sum_{j=1}^K \phi(c_j)$, we have:
\begin{equation}
\delta A_i = -\frac{f(G) \cdot \phi(c_i) - \frac{1}{K}\sum_{j=1}^K f(G) \cdot \phi(c_j)}{s(G)} = -\left( \frac{f(G)}{s(G)} \right) \big( \phi(c_i) - \bar{\phi}_G \big).
\end{equation}
If $f(G)$ is chosen independent of $s(G)$ (e.g., setting $f(G) = \beta$), the ratio exhibits an asymptotic explosion as homogenization occurs:
\begin{equation}
\lim_{s(G) \to 0} |\delta A_i| = \lim_{s(G) \to 0} \frac{\beta |\phi(c_i) - \bar{\phi}_G|}{s(G)} = \infty.
\end{equation}
This induces severe numerical instability and gradient explosion during training. Conversely, choosing $f(G) = \beta \cdot s(G)$ leads to exact algebraic cancellation of the dispersion parameter:
\begin{equation}
\delta A_i = -\frac{\beta \cdot s(G)}{s(G)} \big( \phi(c_i) - \bar{\phi}_G \big) = -\beta \big( \phi(c_i) - \bar{\phi}_G \big).
\end{equation}
Consequently, $\lim_{s(G) \to 0} |\delta A_i| = \beta |\phi(c_i) - \bar{\phi}_G| < \infty$, ensuring uniform stability across all optimization regimes. Setting $\phi(c) \triangleq \varphi(c)$ guarantees scale compatibility.
\end{proof}

\subsection{Linear Contraction of the Category-Mean Channel}
\label{subsec:shrinkage}

We now examine how the established calibration term $\Delta_i = \beta \cdot s(G) \cdot \varphi(c_i)$ modulates the individual components of the advantage decomposition established in Proposition~\ref{prop:advantage_decomp}.

\begin{lemma}[Exact Group-Centering of Standardized Scores]
\label{lem:centering}
Let $\bar{\varphi}_G \triangleq \frac{1}{K} \sum_{j=1}^K \varphi(c_j)$ denote the group average of normalized category deviations. The centered deviation strictly satisfies:
\begin{equation}
\varphi(c_i) - \bar{\varphi}_G = \frac{\mu_{c_i} - \bar{\mu}_G}{\sigma_\mu}.
\end{equation}
\end{lemma}

\begin{proof}
Substituting Equation~\ref{eq:phi_def} into the group mean expression:
\begin{equation}
\bar{\varphi}_G = \frac{1}{K} \sum_{j=1}^K \left( \frac{\mu_{c_j} - \bar{\mu}}{\sigma_\mu} \right) = \frac{1}{\sigma_\mu} \left( \frac{1}{K} \sum_{j=1}^K \mu_{c_j} - \frac{1}{K} \sum_{j=1}^K \bar{\mu} \right) = \frac{\bar{\mu}_G - \bar{\mu}}{\sigma_\mu}.
\end{equation}
Subtracting this from $\varphi(c_i) = (\mu_{c_i} - \bar{\mu}) / \sigma_\mu$, the common constant term $\bar{\mu}$ cancels directly:
\begin{equation}
\varphi(c_i) - \bar{\varphi}_G = \frac{(\mu_{c_i} - \bar{\mu}) - (\bar{\mu}_G - \bar{\mu})}{\sigma_\mu} = \frac{\mu_{c_i} - \bar{\mu} - \bar{\mu}_G + \bar{\mu}}{\sigma_\mu} = \frac{\mu_{c_i} - \bar{\mu}_G}{\sigma_\mu},
\end{equation}
which completes the proof.
\end{proof}

\begin{theorem}[Linear Contraction of the Category-Mean Channel]
\label{thm:shrinkage}
Under the calibrated reward formulation $r_i' = r_i - \beta \cdot s(G) \cdot \varphi(c_i)$, the fixed-dispersion normalized score $\widetilde{A}_i$ satisfies the exact algebraic identity:
\begin{equation}
\widetilde{A}_i = \underbrace{\frac{\delta q_i - \overline{\delta q}_G}{s(G)}}_{A_i^{\mathrm{intra}}} + \left( 1 - \frac{\beta \cdot s(G)}{\sigma_\mu} \right) \underbrace{\frac{\mu_{c_i} - \bar{\mu}_G}{s(G)}}_{A_i^{\mathrm{inter}}}.
\label{eq:thm_shrinkage}
\end{equation}
\end{theorem}

\begin{proof}
Combining the exact decomposition from Proposition~\ref{prop:advantage_decomp} with the net displacement from Proposition~\ref{prop:singularity}, we write:
\begin{equation}
\widetilde{A}_i = A_i + \delta A_i = \left( \frac{\delta q_i - \overline{\delta q}_G}{s(G)} + \frac{\mu_{c_i} - \bar{\mu}_G}{s(G)} \right) - \beta \big( \varphi(c_i) - \bar{\varphi}_G \big).
\end{equation}
Applying Lemma~\ref{lem:centering} to substitute $\varphi(c_i) - \bar{\varphi}_G$:
\begin{equation}
\widetilde{A}_i = \frac{\delta q_i - \overline{\delta q}_G}{s(G)} + \frac{\mu_{c_i} - \bar{\mu}_G}{s(G)} - \frac{\beta}{\sigma_\mu} (\mu_{c_i} - \bar{\mu}_G).
\end{equation}
Factoring out the term $(\mu_{c_i} - \bar{\mu}_G)$ from the second and third components yields:
\begin{equation}
\widetilde{A}_i = \frac{\delta q_i - \overline{\delta q}_G}{s(G)} + \left( \frac{1}{s(G)} - \frac{\beta}{\sigma_\mu} \right) (\mu_{c_i} - \bar{\mu}_G).
\end{equation}
Extracting the factor $1/s(G)$ from the bracketed expression:
\begin{equation}
\left( \frac{1}{s(G)} - \frac{\beta}{\sigma_\mu} \right) (\mu_{c_i} - \bar{\mu}_G) = \left( 1 - \frac{\beta \cdot s(G)}{\sigma_\mu} \right) \left( \frac{\mu_{c_i} - \bar{\mu}_G}{s(G)} \right).
\end{equation}
Substituting this back yields \Eqref{eq:thm_shrinkage}, completing the proof.
\end{proof}

Theorem~\ref{thm:shrinkage} reveals the dual benefits of our calibration operator:
First, the intra-category quality competition channel $A_i^{\mathrm{intra}}$ retains a multiplier strictly equal to unity, ensuring that micro-level relative generation quality is preserved without attenuation.
Second, the category-mean channel $A_i^{\mathrm{inter}}$ is scaled by the contraction factor $\gamma(G) \triangleq 1 - \beta \cdot s(G) / \sigma_\mu$, where $\gamma(G) \in (0, 1)$ under standard hyperparameter configurations. 
Compared to enforcing hard cancellation ($\gamma(G) = 0$), which requires $\beta \propto 1/s(G)$ and risks asymptotic instability as $s(G) \to 0$, soft shrinkage preserves training stability while maintaining meaningful exploration over legitimate category-level quality variations.

%% file: sections/appendix_setup.tex
\section{Experimental Setup}
\label{app:setup}

\paragraph{Overview.}
We evaluate REI on three sources of performance-oriented evolution traces:
an internal industrial corpus from a large-scale production recommender system,
a public benchmark constructed from modded-nanogpt speedrun records,
and a second public benchmark based on rustc-perf
(Appendix~\ref{app:rustc}).
In all three settings, each sample links a pre-change system state, a
performance-motivated modification, and an observed post-change outcome. The full
data construction and evaluation pipelines are programmatic, with no human
intervention in sample selection, split assignment, or metric computation.

\paragraph{Industrial corpus.}
The industrial corpus contains 60,000+ recommender-system evolution samples,
covering success/failure annotations of changes from 24 code repositories.
Among them, 54,710 samples are used for training and 6,291 for testing.

Each raw industrial record corresponds to one recommender-system evolution attempt
and contains three parts:
(1) an engineering-code snapshot, including the full training repository at
submission time;
(2) training logs and metrics, including loss curves and downstream evaluation
outputs; and
(3) experiment metadata, such as repository identifiers, commit hash,
experiment id, timestamps, runner, model name, dataset identifier, and
checkpoint path.
This structure matches the real evolution loop in production:
a candidate direction is first implemented, then trained and evaluated,
and only afterwards can its value be determined.

\paragraph{Industrial data construction.}
To reduce sparsity, we form pairwise samples from historical experiment trajectories
rather than keeping only strictly adjacent changes. To reduce noise, we apply a
fully automatic two-stage filtering pipeline.

The first stage is execution-logic filtering. We perform reachability analysis from
the script entry point and discard changed functions or classes that are not
reachable from actual execution paths. We then parse Python files into abstract
syntax trees (ASTs), remove debugging statements and comments, normalize import
ordering, and suppress purely superficial edits. After normalization, we compute
pairwise code differences and apply configuration whitelisting so that engineering
variables such as output directories, resource settings, tracing options, and data
endpoints are removed, while performance-relevant fields such as optimizer,
learning rate, weight decay, batch size, model structure, and feature configuration
are retained.

The second stage is semantic intent filtering. The purified code delta is mapped
into explicit evolution intent by a fixed extraction pipeline. Modifications whose
intent cannot be reliably recovered are discarded as ambiguous noise. The retained
samples are further organized into structured direction annotations used by the
generator and by the diversity analysis.

\paragraph{NanoGPT benchmark.}
The nanoGPT benchmark is built from the modded-nanogpt speedrun leaderboard
(\texttt{KellerJordan/modded-nanogpt}), using pull requests and associated
repository history from 2024-05 to 2026-01. We use two leaderboard tracks:
track~1 (GPT-2 Small) and track~2 (GPT-2 Medium). We choose this source because
it naturally records repeated performance-oriented iteration: contributors propose
a change, realize it in code, run the full training/evaluation pipeline, and
report the resulting gain. Compared with generic open-source repository histories,
these records are cleaner for ideation learning because the dominant intent is
leaderboard improvement rather than maintenance, refactoring, or feature
development.

Each retained public item is annotated with one of five direction families: C1 system/kernel, C2 optimizer, C3 architecture, C4 attention, and C5 loss. On the public benchmark, these categories are the ground-truth direction families provided by the dataset and are used for method-level evaluation.

\paragraph{NanoGPT data curation and split.}
The raw nanoGPT library contains 190 candidate records. After removing invalid entries, we retain 107 valid ideas for the benchmark. Retained items are labeled by observed outcome: \emph{strong} for realized speedup $\geq 1\%$ and \emph{null} for valid attempts without meaningful gain. Invalid records are excluded from the curated set. We train on 79 track-1 items and evaluate on a 28-question held-out test set, of which 23 have ground-truth labels for family-level evaluation. The split is defined at the PR level to prevent leakage, so all artifacts associated with a held-out PR are kept out of training.

\paragraph{Prompt construction.}
Each prompt contains the current scenario context together with up to 8 historical
changes retrieved from the corresponding training library. The model is asked to
generate a compact single-line proposal in the format
\texttt{[CATEGORY] <what to change>}.

In the public benchmark, the ground-truth label is the accepted direction family
attached to the held-out record. In the industrial setting, where no ground-truth taxonomy exists, family labels used for diversity analysis are obtained by the same LLM extraction pipeline used during industrial data construction.

\paragraph{Class balance.}
To reduce trivial under-exposure of rare families, SFT up-samples minority
categories in the public training set. We additionally tested whether the observed
semantic collapse could be explained by family-frequency skew in the data or reward
pipeline. It could not: the collapse remains qualitatively unchanged after category
rebalancing, indicating that the phenomenon is not an artifact of popularity bias alone.

\paragraph{Implementation.}
The nanoGPT benchmark is relatively small, so nanoGPT-benchmark experiments use
\texttt{Qwen2.5-0.5B} as the default backbone and run on 2 NVIDIA A100 80GB GPUs
with standard data parallelism. By contrast, the industrial corpus is much larger
and the industrial experiments use \texttt{Qwen3-8B} on 8 NVIDIA A100 80GB GPUs.
Unless otherwise stated, SFT, reward-model training, and GRPO follow the same
stage definitions described in the main paper, with hardware scaled to match the
dataset size and backbone.

Rollout sampling uses temperature $0.99$, top-$p=0.99$,
and top-$k=100$. Megatron parallelism is disabled beyond data parallelism
(TP=PP=EP=1). Software versions are PyTorch 2.6.0, vLLM 0.8.4, and
Transformer-Engine 2.2.0.

The main-text nanoGPT-benchmark tables report a single representative seed
(\(42\)); five-seed means over \(\{42, 211, 985, 43, 44\}\) and paired
significance tests are given in Appendix~\ref{app:multiseed}. All post-training
baselines share the same SFT
initialization, reward model, training prompts, and test set, and differ only in
the RL objective or diversity-preserving mechanism.

\paragraph{Evaluation protocol.}
For each test prompt, the model generates 8 candidates unless otherwise specified, ranked by the corresponding reward model. Candidate-set quality uses four metrics. Hit@1 is the fraction of prompts whose top-ranked candidate falls in a correct method family. Hit@8 is the fraction for which at least one of the 8 candidates falls in a correct family. Coverage is, for prompts with multiple valid families, the fraction of those families covered by the candidate set. TV to Reference Prior is the total variation distance between the generated method-category distribution and the reference-corpus prior, where smaller is better. In the industrial setting, each evaluation instance has a single ground-truth family, so Coverage is identical to Hit@8. In the public benchmark, a prompt may have multiple valid families because several distinct PR directions from the same repository state are treated as valid targets, so Coverage and Hit@8 need not coincide.

Method-level diversity uses four metrics. Distinct Method Count is the number of distinct families among the 8 candidates. Top-Family Share is the share of the most frequent family. Max Within-Family Count is the largest number of candidates in a single family. Single-Family Collapse Rate is the fraction of prompts whose 8 candidates all fall in one family. Cross-task evaluates transfer beyond the training task, while long-tail removes the dominant head family before recomputing the metrics, testing whether a method can still recover valid non-head families.

\paragraph{Fairness of comparison.}
All compared post-training baselines are trained from the same SFT checkpoint and evaluated on the same held-out prompts with the same candidate budget. For industrial offline evaluation, the same system context and candidate budget are given to all model baselines. All model outputs are processed under the same ranking and metric pipeline, with no manual intervention or example-specific adjustment. Human-expert comparisons use the same task description and access to the same scenario context. The method-category signal is provided to all compared methods as a shared input format. Baselines receive the same [CATEGORY]-tagged prompts but retain their original objectives. SPOD differs in that it uses this shared signal for category-aware reward correction and top-1 retention.


%% file: sections/appendix_analysis_protocol.tex
\section{Protocols for the Diversity-Collapse Analysis}
\label{app:analysis_protocol}

\paragraph{Evaluation data and slices.}
Figures~\ref{fig:analysis} and~\ref{fig:surface_method} use the public
nanoGPT-speedrun benchmark with the 79-item training split and 28-prompt
held-out test split, with Qwen2.5-0.5B as the ideation policy. Ten late-stage
prompts from track~1 form the same-task slice and evaluate temporal
extrapolation within the training track. The remaining 18 prompts come from
track~2 and form the cross-task slice, evaluating transfer to a track unseen
during training. Both slices are fully held out from policy training.

\paragraph{Reward and long-tail summaries.}
For each prompt, the policy generates eight candidates, ranked by the same
verifier used during training. In Figure~\ref{fig:analysis}(a), mean reward is
computed by averaging verifier scores over the eight candidates and then over
prompts. Best-of-8 reward first takes the highest verifier score among the
eight candidates for each prompt and then averages across prompts. The former
summarizes overall candidate-set quality, whereas the latter measures whether
sampling still produces at least one high-scoring direction.

For Figure~\ref{fig:analysis}(b), generated candidates are mapped to the five
speedrun method families: system/kernel, optimizer, architecture, attention,
and loss. Families are grouped by generated share into head ($\geq 40\%$), mid
($10\%$--$40\%$), and tail ($<10\%$), after which Hit@1 is recomputed within
each bucket. Prompts without a usable ground-truth family label are excluded
from Hit@1. This analysis tests whether post-training gains are broadly
distributed or concentrated in a small number of dominant families.

\paragraph{Surface- and method-level diversity.}
Figure~\ref{fig:surface_method} separates surface-level from method-level
changes. The left panel reports token-level entropy during generation. The
middle panel tracks three method-level statistics normalized to step~0:
category entropy, the share of the most frequent family, and the number of
distinct families $n_{\mathrm{cat}}$ covered by the eight candidates. The
right panel reports final-checkpoint percentage changes relative to the SFT
initialization. Surface statistics are computed directly from generated text:
word-level Distinct-2 measures bigram repetition, the TF--IDF Vendi score
measures the effective number of distinct texts, and mean pairwise cosine
distance measures separation in TF--IDF space. Method-level statistics use
the offline family annotations. All percentage changes use the same
$28\times 8$ evaluation batch for the method being compared with its SFT
initialization.

\paragraph{Verifier-fidelity intervention.}
Figure~\ref{fig:verifier_fidelity} studies whether stronger verifier agreement
is associated with greater method-level concentration. Verifier fidelity is
defined as pairwise ranking accuracy between verifier scores and realized
gains, where $0.5$ denotes random ranking and $1.0$ perfect agreement. Realized
gains come from the nanoGPT-speedrun submission history. For each submission,
we compute the relative wall-time change from the adjacent record and apply
same-era quantile normalization to reduce temporal drift. This reduces the
Spearman correlation between record index and normalized gain from $-0.46$ to
$-0.08$. Historical submissions are grouped into strong, null, and reject
outcomes using the community significance threshold $p<0.01$.

To vary fidelity while holding the remaining setup fixed, we keep the deployed
verifier, policy initialization, training data, model scale, and optimization
configuration unchanged and perturb only the verifier score:
\begin{equation}
  s_{\lambda}=s_{\mathrm{RM}}+\lambda\epsilon,
  \qquad \epsilon\sim\mathcal{N}(0,1),
\end{equation}
where $\lambda$ controls the noise magnitude. The weak, medium, and strong
conditions have measured pairwise ranking accuracies of $0.55$, $0.67$, and
$0.83$, respectively. After training under each condition, concentration is
measured as the final share of generated candidates assigned to the most
frequent method family. The corresponding shares are $49.1\%$, $58.8\%$, and
$62.5\%$.

%% file: sections/appendix_baselines.tex
\section{Baseline Methods}
\label{app:baselines}

We compare SPOD with supervised learning, reward-driven policy optimization, and token-level exploration controls. For a prompt $x$, let $y_i$ be a sampled response, $s_{i,t}=(x,y_{i,<t})$ its prefix, $R_i$ its reward, and $\rho_{i,t}=\pi_\theta(y_{i,t}\mid s_{i,t})/\pi_{\mathrm{old}}(y_{i,t}\mid s_{i,t})$ the importance ratio. Write $H_{i,t}$ for the entropy of the full next-token distribution at that prefix.

\paragraph{Supervised fine-tuning (SFT).}
SFT maximizes the likelihood of reference responses, minimizing $-\mathbb{E}_{(x,y)\sim\mathcal D}\log\pi_\theta(y\mid x)$. In REI, repository-derived examples provide supervision for learning reusable ideation patterns. SFT provides a reference before reward optimization, without explicitly optimizing candidate-set coverage.

\paragraph{Group Relative Policy Optimization (GRPO).}
GRPO generates multiple responses per prompt and computes $A_i=(R_i-\bar R)/(\operatorname{std}(R)+\epsilon)$ within each group, avoiding a separate value model. Its clipped policy surrogate averages $\min\{\rho_{i,t}A_i,\operatorname{clip}(\rho_{i,t},1-\epsilon_l,1+\epsilon_h)A_i\}$ over tokens and responses. It reinforces candidates with above-average rewards.

\paragraph{Entropy regularization (Ent. Reg.).}
This baseline augments reward optimization with an entropy bonus, $J_{\mathrm{RL}}+\alpha\mathbb{E}_{i,t}[H_{i,t}]$, where $\alpha>0$ controls the incentive to retain uncertainty. It discourages premature concentration of next-token probabilities. However, preserving token entropy does not directly ensure that generated ideas span distinct methods: different wordings can still express the same solution family.

\paragraph{Entropy-based advantage adjustment (Entro. Adv.).}
These variants incorporate conditional entropy into the advantage or its weighting, making policy updates depend on both reward feedback and uncertainty at a generation step. Unlike an additive entropy regularizer, they modify the learning signal attached to sampled tokens. Such interventions aim to preserve exploration at uncertain positions, but uncertainty alone does not determine whether a particular update increases or decreases entropy.

\paragraph{Clipped IS-weight Policy Optimization (CISPO).}
CISPO clips importance weights while retaining a log-likelihood gradient:
\begin{equation*}
J_{\mathrm{CISPO}}=\mathbb E\!\left[\frac{1}{\sum_i |y_i|}\sum_{i,t}
\operatorname{sg}\!\left[\operatorname{clip}(\rho_{i,t},1-\epsilon_l^{\mathrm{IS}},1+\epsilon_h^{\mathrm{IS}})\right]
A_i\log\pi_\theta(y_{i,t}\mid s_{i,t})\right],
\end{equation*}
where $\operatorname{sg}$ stops gradients through the weight. Unlike the clipped GRPO surrogate, exceeding a ratio threshold does not itself eliminate the token's direct policy gradient. This preserves learning signals while limiting their importance weights; it does not recalibrate the underlying reward or identify redundant method families.

\paragraph{STEER.}
STEER estimates the local entropy change induced by a policy update from token probabilities, advantages, and conditional entropy. It attenuates updates with large predicted changes using $\lambda_{i,t}=\exp(-k|\Omega_{i,t}|)$, where $\Omega_{i,t}$ estimates entropy change and $k$ sets the minimum weight. Thus it regulates the rate of entropy change, rather than simply maximizing entropy, and suppresses both abrupt concentration and excessive dispersion.

\paragraph{Decoupled Clip and Dynamic Sampling Policy Optimization (DAPO).}
DAPO combines asymmetric ratio clipping, dynamic sampling of informative response groups, token-level loss aggregation, and overlong-response reward shaping. A higher upper clipping threshold gives positively rewarded, initially unlikely tokens more room to grow; dynamic sampling excludes groups with identical rewards and no relative learning signal. These baselines test whether generic exploration controls suffice to preserve method diversity, compared with SPOD's category-aware reward correction.

%% file: sections/appendix_sensitivity.tex
\section{Parameter Sensitivity}
\label{app:sensitivity}

\begin{figure}[htbp]
  \centering
  \includegraphics[width=0.8\textwidth]{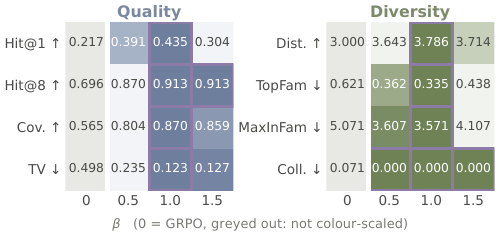}
  \caption{Sensitivity to the debiasing strength $\beta$. Abbreviations follow
  Table~\ref{tab:ablation_spod}. The $\beta=0$ (GRPO) column is greyed out and
  excluded from the per-row min--max scaling; raw values are shown and the
  best value in each row is outlined.}
  \label{fig:beta}
\end{figure}

\paragraph{Debiasing strength.}
The debiasing strength $\beta$ controls the reward correction, with $\beta=0$ reducing exactly to GRPO. We sweep $\beta\in\{0,0.5,1.0,1.5\}$ and find that SPOD is robust across a broad range of values: all nonzero choices of $\beta$ outperform the $\beta=0$ GRPO baseline overall across the eight metrics in Figure~\ref{fig:beta}. Among them, $\beta=1.0$ gives the best overall trade-off and is therefore used in all other experiments. Increasing $\beta$ further to $1.5$ causes some metrics to regress, suggesting that overly strong correction can start to suppress genuinely strong in-family candidates, although it still remains better overall than GRPO.

\begin{figure}[htbp]
  \centering
  \includegraphics[width=0.72\textwidth]{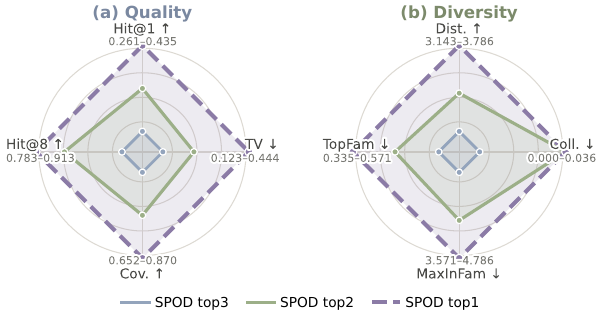}
  \caption{Sensitivity to top-$k$ exemption. Axes are min--max scaled across the tested settings; outward is better. Metric abbreviations follow Table~\ref{tab:ablation_spod}.}
  \label{fig:topk}
\end{figure}

\paragraph{Exemption depth.}
We further ablate the exemption depth $k$, i.e.\ how many top-ranked in-family
candidates are shielded from the penalty. Tightening $k$ from $3$ to $1$
improves all eight metrics monotonically (Figure~\ref{fig:topk}): a
larger $k$ shields more of the in-family head and thus dilutes the debiasing
signal, while $k=1$ is the tightest setting that still protects the strongest
in-family candidate. We use $k=1$ in the remaining experiments.

\paragraph{Default setting across datasets.}
Unless otherwise stated, all reported experiments on the industrial system,
nanoGPT, and rustc-perf use the same SPOD hyperparameters, namely
$\beta=1.0$ and $k=1$. We do not tune these parameters separately for each
dataset.

%% file: sections/appendix_industrial_slices.tex
\section{Industrial Cross-Task and Long-Tail Evaluation}
\label{app:industrial_slices}

\paragraph{Evaluation scope.}
We report additional industrial offline evaluations on two slices: cross-task, which assesses generalization beyond the training task, and long-tail, which evaluates performance on long-tail method categories. Table~\ref{tab:industrial_cross_longtail} reports the four quality metrics for all eight methods. Each industrial instance has a single ground-truth method category, so Coverage equals Hit@8 for every method: covering the ground truth is equivalent to having at least one matching candidate.  Bold entries indicate column-wise best values.

\begin{table}[htbp]
\centering
\normalsize
\caption{Industrial offline quality on cross-task and long-tail slices. Coverage equals Hit@8 because each instance has one ground-truth category. Higher is better except for TV to Reference Prior.}
\label{tab:industrial_cross_longtail}
\begin{tabular*}{\textwidth}{@{\extracolsep{\fill}}lcccccccc}
\toprule
\multirow{2}{*}{Method} & \multicolumn{4}{c}{Cross-Task} & \multicolumn{4}{c}{Long-Tail} \\
\cmidrule(lr){2-5} \cmidrule(lr){6-9}
& Hit@1$\uparrow$ & Hit@8$\uparrow$ & Cov.$\uparrow$ & TV$\downarrow$
& Hit@1$\uparrow$ & Hit@8$\uparrow$ & Cov.$\uparrow$ & TV$\downarrow$ \\
\midrule
SFT & 0.500 & 0.885 & 0.885 & 0.112 & 0.248 & 0.426 & 0.426 & 0.093 \\
GRPO & 0.495 & 0.825 & 0.825 & \textbf{0.069} & 0.278 & 0.422 & 0.422 & 0.087 \\
STEER & 0.500 & 0.890 & 0.890 & 0.104 & 0.300 & 0.440 & 0.440 & 0.075 \\
DAPO & 0.515 & 0.900 & 0.900 & 0.229 & 0.314 & 0.441 & 0.441 & 0.168 \\
Ent. Reg. & 0.490 & 0.890 & 0.890 & 0.117 & 0.266 & 0.429 & 0.429 & 0.209 \\
CISPO & 0.475 & 0.849 & 0.849 & 0.203 & 0.255 & 0.502 & 0.502 & 0.191 \\
Entro. Adv. & 0.519 & 0.882 & 0.882 & 0.303 & 0.313 & 0.497 & 0.497 & \textbf{0.068} \\
SPOD & \textbf{0.579} & \textbf{0.916} & \textbf{0.916} & 0.085 & \textbf{0.387} & \textbf{0.570} & \textbf{0.570} & 0.080 \\
\bottomrule
\end{tabular*}
\end{table}

\paragraph{Cross-task results.}
SPOD achieves the highest Hit@1 (0.579), Hit@8 (0.916), and Coverage (0.916). Relative to the best competing baseline for each metric, its Hit@1 exceeds Entro. Adv. by 0.060 (6.0 percentage points), while its Hit@8 and Coverage exceed DAPO by 0.016 (1.6 percentage points). 

\paragraph{Long-tail results.}
SPOD again attains the highest Hit@1 (0.387), Hit@8 (0.570), and Coverage (0.570). Its Hit@1 exceeds DAPO, the strongest alternative on this metric, by 0.073 (7.3 percentage points); its Hit@8 and Coverage exceed CISPO by 0.068 (6.8 percentage points). For TV, Entro. Adv. ranks first (0.068), followed by STEER (0.075) and SPOD (0.080). These results support improved recovery of valid long-tail methods, while indicating that candidate-level accuracy and agreement with the reference category distribution capture different aspects of performance.

%% file: sections/appendix_rustc.tex
\section{Generalization to a Second Public Benchmark: rustc-perf}
\label{app:rustc}

The main paper evaluates SPOD in two application scenarios: an industrial recommender system (\emph{AI for Recommendation}) and the nanoGPT speedrun (\emph{AI for AI}). To test whether the quality--diversity behavior of SPOD transfers to a qualitatively different scenario, we evaluate SPOD in a third setting---\emph{AI for Software Engineering}---using a second, fully public benchmark drawn from the Rust compiler ecosystem. In this task, given the recent optimization history of the compiler, the model proposes one next change that may reduce compilation time on the rustc-perf benchmark suite. The corpus contains $1{,}591$ landed changes collected from public rustc-perf triage reports and rust-lang/rust pull requests, and a strict temporal split yields $1{,}414$ training states and $177$ held-out test states. At each state the model observes summary statistics from the preceding $52$ weeks and the eight most recent compiler changes, then proposes one concrete optimization in the form ``\texttt{[category] change description}''. The six categories cover backend and code generation, type and trait solving, incremental queries, frontend parsing, the standard library, and infrastructure and tooling. Crucially, the gold utility is derived from the compile-time change measured by the public rustc-perf continuous-integration reports rather than from any model-generated judgment, so the evaluation does not rely on a private task or a learned gold signal.

\begin{table}[htbp]
\centering
\caption{Summary of the public rustc-perf benchmark.}
\label{tab:rustc_data}
\begin{tabular}{ll}
\toprule
Item & Setting \\
\midrule
Source & Public rustc-perf triage reports and rust-lang/rust pull requests \\
Base model & Qwen3-1.7B-Base \\
Task granularity & One state per landed compiler change \\
Dataset size & $1{,}591$ states: $1{,}414$ train / $177$ temporal hold-out \\
Evaluation & $k=8$ samples per state; $157$ of $177$ test states carry usable gold labels \\
Gold signal & Measured compile-time change from rustc-perf CI \\
\bottomrule
\end{tabular}
\end{table}

We report the same eight metrics as in the main paper, which separate answer quality from category-level diversity. Hit@1, Hit@8, Coverage, and the Distinct Method Count (Dist.) are better when larger, while TV to the reference category distribution, Top-Family Share (TopFam), Max Within-Family Count (MaxInFam), and the Single-Family Collapse Rate (Coll.) are better when smaller. All methods are evaluated on the same temporal hold-out with the same reward model. Tables~\ref{tab:rustc_quality} and~\ref{tab:rustc_diversity} report quality and diversity, respectively; bold marks the best result in each column.

\begin{table}[htbp]
\centering
\caption{Quality results on the rustc-perf benchmark.}
\label{tab:rustc_quality}
\begin{tabular*}{\textwidth}{@{\extracolsep{\fill}}lcccc}
\toprule
Method & Hit@1$\uparrow$ & Hit@8$\uparrow$ & Cov.$\uparrow$ & TV$\downarrow$ \\
\midrule
SFT & 0.236 & 0.803 & 0.727 & 0.226 \\
GRPO & 0.198 & 0.809 & 0.727 & 0.252 \\
DAPO & 0.229 & 0.764 & 0.702 & 0.270 \\
Entro.\ Adv. & 0.223 & 0.783 & 0.706 & 0.232 \\
Ent.\ Reg. & 0.210 & 0.803 & 0.732 & 0.235 \\
CISPO & 0.248 & 0.847 & 0.770 & 0.247 \\
STEER & 0.248 & 0.777 & 0.715 & 0.240 \\
\midrule
SPOD & \textbf{0.261} & \textbf{0.860} & \textbf{0.773} & \textbf{0.218} \\
\bottomrule
\end{tabular*}
\end{table}

\begin{table}[htbp]
\centering
\caption{Category-diversity results on the rustc-perf benchmark. Coll.\ is the fraction of states for which all eight samples fall into one category.}
\label{tab:rustc_diversity}
\begin{tabular*}{\textwidth}{@{\extracolsep{\fill}}lcccc}
\toprule
Method & Dist.$\uparrow$ & TopFam$\downarrow$ & MaxInFam$\downarrow$ & Coll.$\downarrow$ \\
\midrule
SFT & \textbf{4.616} & 0.233 & \textbf{2.989} & \textbf{0.000} \\
GRPO & 4.418 & 0.262 & 3.057 & \textbf{0.000} \\
DAPO & 4.463 & 0.260 & 3.096 & \textbf{0.000} \\
Entro.\ Adv. & 4.362 & 0.258 & 3.085 & \textbf{0.000} \\
Ent.\ Reg. & 4.514 & 0.259 & 3.209 & \textbf{0.000} \\
CISPO & 4.463 & 0.254 & 3.040 & \textbf{0.000} \\
STEER & 4.412 & 0.253 & 3.085 & \textbf{0.000} \\
\midrule
SPOD & 4.537 & \textbf{0.201} & 3.051 & \textbf{0.000} \\
\bottomrule
\end{tabular*}
\end{table}

\paragraph{Main finding.}
On this public compiler-optimization task, standard reinforcement learning increases category concentration relative to the SFT initialization: GRPO raises Top-Family Share from $0.233$ to $0.262$ while reducing the Distinct Method Count. SPOD reverses this trend, attaining the lowest Top-Family Share ($0.201$) together with the highest Hit@1 ($0.261$), the highest Hit@8 ($0.860$), the highest Coverage ($0.773$), and the lowest TV ($0.218$). These results reproduce the central quality--diversity claim of the main paper on an independent public benchmark whose gold signal comes from measured compile-time changes rather than from model-generated judgments.

%% file: sections/appendix_replay.tex
\section{Multi-Round Replay Simulation: Details}
\label{app:rsi}

This appendix expands the replay simulation summarized in Section~\ref{sec:multi_round}. The core issue is not diversity in the abstract, but a concrete optimization behavior: under reward-only training, policies increasingly shift probability mass toward the head method family that is easiest to score well under the current reward model or dataset prior. This can preserve superficial single-round performance, but over repeated iteration it systematically removes non-head yet genuinely high-yield directions. We observe the same overall pattern on both the public speedrun data and the industrial dataset; Section~\ref{app:rsi_ind} reports the industrial replication.

\subsection{Public Benchmark: Family-Level Concentration and Long-Horizon Stagnation}

On the public speedrun data, we sort the historically accepted improvements within each method family by realized gain. At each round, a policy samples one family according to the trained model's measured output distribution and receives the next unused real gain from that family. This setup asks a direct question: if a policy keeps most of its probability mass on a small number of families, does repeated reuse of the same high-prior direction eventually hurt long-horizon performance?

\begin{figure}[htbp]
\centering
\includegraphics[width=\linewidth]{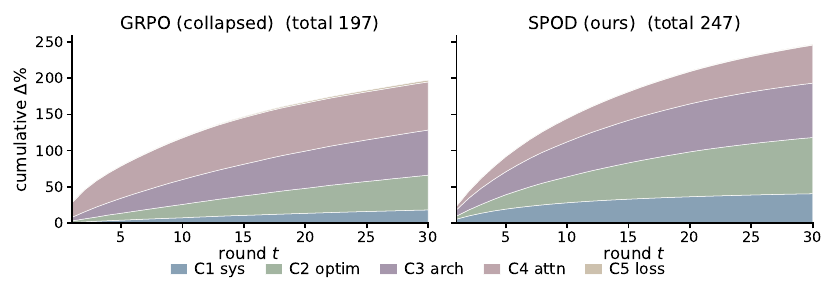}
\caption{Family-decomposed cumulative reward on the public benchmark at $T{=}30$.
GRPO (left) derives most of its reward from a single dominant family and grows
more slowly later in the horizon. SPOD (right), by contrast, maintains broader
coverage across C1--C5 and continues to accumulate reward from multiple
families throughout the horizon.}
\label{fig:rsi-depth}
\end{figure}

The replay results support this concern. GRPO concentrates heavily on the head
family C4, allowing it to obtain some early reward from attention-related
directions. However, its cumulative reward grows more slowly in later rounds.
Under $T{=}30$, GRPO spends on average 8.96 rounds on families that no longer
yield additional replay reward. By contrast, SPOD preserves broader coverage
across C1--C5, so its replay distribution continues to draw reward from multiple
families in later rounds (Figure~\ref{fig:rsi-depth}). This pattern is
consistent with SPOD's advantage over GRPO under short-horizon, long-horizon,
and cross-task settings, and with the widening gap as the horizon grows.

\begin{figure}[htbp]
\centering
\includegraphics[width=\linewidth]{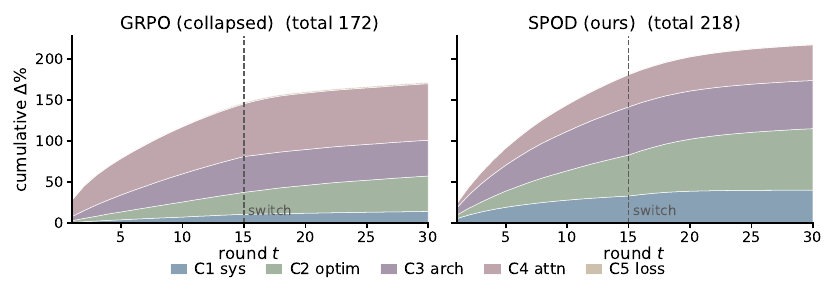}
\caption{Cross-task replay on the nanoGPT benchmark. The task pool switches at
round~15 (dashed line). GRPO remains concentrated on its previous head-family
preference, whereas SPOD maintains broader coverage and retains probability mass
on families that continue to yield reward after the switch.}
\label{fig:rsi-cross}
\end{figure}

The cross-task result supports the same interpretation. After the task pool
switches at round~15, GRPO remains more concentrated on its previous
head-family preference. Because SPOD preserves broader family coverage, its
replay distribution continues to place probability mass on families that remain
productive in the new task pool, allowing it to stay ahead after the switch
(Figure~\ref{fig:rsi-cross}).

\subsection{Industrial Replication: Misalignment Between Popularity and Productivity}
\label{app:rsi_ind}

The industrial replay leads to the same conclusion, but the key pattern is a systematic misalignment between popularity and productivity. In the industrial library, the most common method family is not the family whose individual directions contribute the strongest returns. However, reward-only RL continues to push probability mass toward the most common and easiest-to-repeat family, rather than the families that contribute the largest long-horizon payoff.

\begin{figure}[htbp]
\centering
\includegraphics[width=\linewidth]{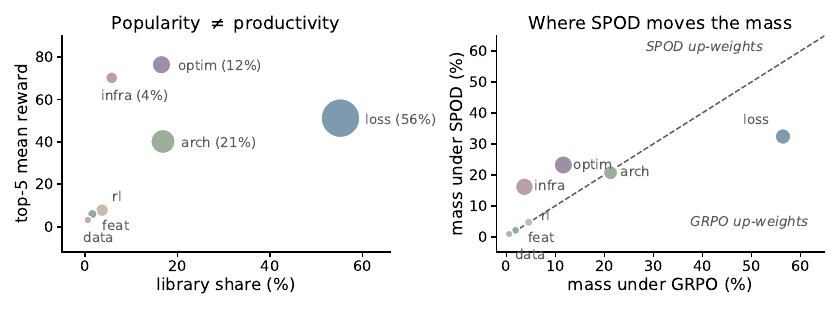}
\caption{Popularity--productivity misalignment on the industrial library. Left:
family popularity (share of historical library) versus productivity (mean realized
gain); the most frequent family is not the most productive one. Right: the family
allocation GRPO learns versus SPOD; marker area encodes GRPO's probability mass on
that family and the largest-mass families are annotated, and markers above the dashed
diagonal are up-weighted by SPOD relative to GRPO. SPOD shifts mass from the popular
head family toward the higher-yield optimization and training-infra families.}
\label{fig:rsi-astra-mech}
\end{figure}

This pattern is clear in the industrial replay (Figure~\ref{fig:rsi-astra-mech}). Loss design is the most frequent family in the historical library, and GRPO further increases its allocation to that family after RL. But the highest-payoff directions are concentrated more heavily in colder families such as optimization and training infrastructure. In other words, GRPO learns that repeatedly betting on the popular family is the easiest way to maintain high reward, not that this family contains the best directions. SPOD acts exactly on this failure mode: it weakens the one-sided concentration toward the head family and reallocates more mass toward optimization and training-infra, which leads to substantially higher cumulative reward over time.

\begin{figure}[htbp]
\centering
\includegraphics[width=\linewidth]{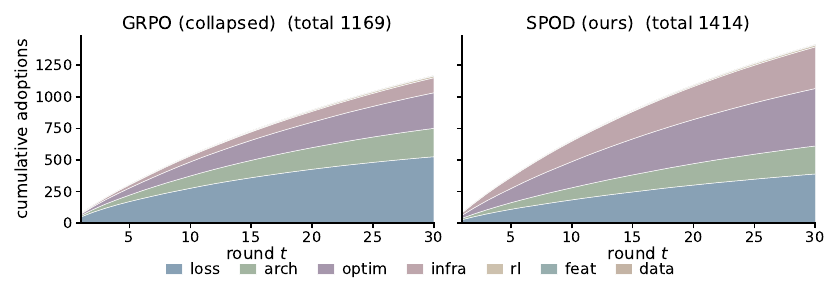}
\caption{Cumulative reward over the horizon on the industrial replay. SPOD stays
above GRPO throughout the horizon. This advantage is consistent with a more
balanced allocation across method families in the replay.}
\label{fig:rsi-astra-cum}
\end{figure}

The industrial results are fully consistent with this mechanism. SPOD outperforms GRPO under short-horizon, long-horizon, and cross-task settings, and comes much closer to the oracle upper bound in the long-horizon setting. This gain is not driven by a few isolated lucky samples. It comes from a persistent difference in family allocation: GRPO obtains too much of its cumulative reward from the head family, while SPOD recovers substantially more useful directions from optimization and training-infra, yielding both higher average reward per round and higher final cumulative reward (Figure~\ref{fig:rsi-astra-cum}).

\begin{figure}[htbp]
\centering
\includegraphics[width=\linewidth]{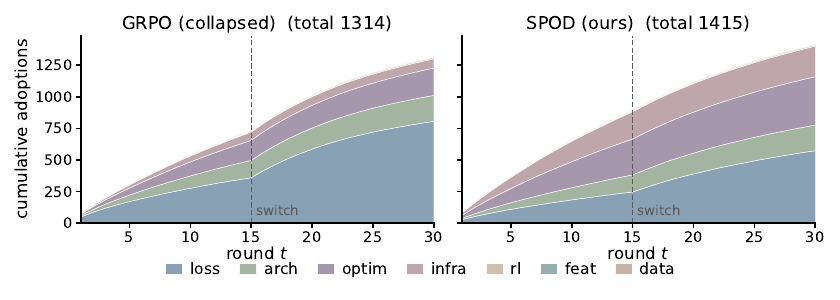}
\caption{Cross-task replay on the industrial dataset. After the task pool switches, SPOD remains ahead and continues to accumulate reward from a broader mix of method families. By contrast, GRPO remains heavily concentrated on the loss family, and most of its post-switch gains still come from loss, with little contribution from other families.}
\label{fig:rsi-astra-cross}
\end{figure}

The cross-task result supports the same explanation. After the task pool switches, both methods encounter new high-payoff directions, but SPOD retains its advantage because its higher allocation to optimization and training-infra continues to transfer to the new task (Figure~\ref{fig:rsi-astra-cross}). Its gain therefore does not depend on any one fixed family, but on a healthier long-horizon allocation policy.

%% file: sections/appendix_cases.tex
\section{Case Studies: Why the Multi-Round Gap Emerges}
\label{app:case}

This section uses four representative examples to explain why the multi-round gap above arises. Across all four cases, the same mechanism appears: after RL, GRPO, STEER, and DAPO increasingly revolve around the head family C4 (attention), because repeatedly emitting that family is the easiest way to align with the current reward-model preference or dataset prior. In the short term, this can preserve or even inflate hit-based scores. In the long term, it systematically removes non-head but genuinely high-yield families from the candidate set. SPOD does not help by enforcing a uniform distribution. It helps by suppressing this head-family bias, so that useful non-head families remain available before multi-round iteration even begins.

\subsection{Case 1: All Three RL Baselines Collapse to the Same Family}

This example is the cleanest illustration of semantic collapse after RL. The prompt comes from track 2 under the cross-track slice, at a relatively early state ($27.7$ minutes, $n\_history=3$). The ground-truth accepted direction is a C2 optimizer change: extending the learning-rate cooldown phase, with a gain of $1.805$ minutes and a strong-tier label.

The key observation is that SFT does \emph{not} hit the ground truth, but it still retains multiple families. After RL, however, all three RL baselines collapse completely to the head family C4. GRPO, STEER, and DAPO each produce 8/8 C4 candidates. At that point, the problem is no longer only ranking error: the correct family has disappeared from the candidate set entirely. Under the same RL setting, SPOD still preserves four families and includes a C2 candidate (Table~\ref{tab:case1}). Its top-ranked candidate is still wrong, but the correct family remains alive in the pool.

\begin{table}[htbp]
\centering
\normalsize
\caption{Case 1: the correct family disappears after RL. The ground truth is a C2 optimizer change, but all three RL baselines collapse to 8/8 C4 candidates.}
\label{tab:case1}
\begin{tabular*}{\textwidth}{@{\extracolsep{\fill}}llc}
\toprule
Arm & Family distribution over 8 candidates & Hit@1 \\
\midrule
SPOD ($\beta{=}1.0$) & C1$\times$3 / C2$\times$1 / C3$\times$2 / C4$\times$2 & miss \\
GRPO & C4$\times$8 & miss \\
STEER & C4$\times$8 & miss \\
DAPO & C4$\times$8 & miss \\
SFT & C1$\times$4 / C3$\times$2 / C4$\times$2 & miss \\
\bottomrule
\end{tabular*}
\end{table}

This case directly explains the long-horizon gap in the replay. A collapsed policy does not merely rank the wrong candidate first; it removes the correct family before ranking even starts. Once this happens repeatedly across prompts, non-head families such as C2 contribute less and less over time. SPOD's first gain is therefore not simply better top-1 ranking, but preventing RL from collapsing the candidate space so aggressively that the correct family disappears.

\subsection{Case 2: The Baselines Keep Betting on Attention While SPOD Preserves the Optimizer Direction}

The second example shows that the same failure mode is not an isolated extreme case, but a stable search bias. This prompt also comes from track 2 under the cross-track slice ($23.4$ minutes, $n\_history=8$). The ground-truth accepted direction is again in C2: the Snoo optimizer, with gain $0.725$.

SPOD produces a mixed candidate set and hits the ground truth at top-1, whereas GRPO, STEER, DAPO, and SFT all allocate five of their eight candidates to C4 and miss (Table~\ref{tab:case2}). More importantly, this is not a matter of guessing the exact method name by chance. The top SPOD candidate stays within the optimizer / learning-rate family, while the baselines' top candidates continue to revolve around attention windows, attention skip connections, and qkv-related changes.

\begin{table}[htbp]
\centering
\normalsize
\caption{Case 2: SPOD preserves the optimizer family, while the baselines continue to bet on attention.}
\label{tab:case2}
\setlength{\tabcolsep}{3pt}
\begin{tabularx}{\textwidth}{@{}>{\raggedright\arraybackslash}p{0.17\textwidth}>{\raggedright\arraybackslash}p{0.24\textwidth}c>{\raggedright\arraybackslash}X@{}}
\toprule
Arm & Family distribution & Hit@1 & First line of top-scored candidate \\
\midrule
SPOD ($\beta{=}1.0$) & C1$\times$2 / C2$\times$1 / C3$\times$4 / C4$\times$1 & \checkmark & \texttt{[C2] Faster step, reduced lr decay duration, reduced learning rate value range, improved parameter freeze \& warmstart logic} \\
GRPO & C3$\times$1 / C4$\times$5 / C5$\times$2 & miss & \texttt{[C4] Attention window update, 1024x residual connection, no gate layer, 1024x qkv layer ...} \\
STEER & C3$\times$1 / C4$\times$5 / C5$\times$2 & miss & \texttt{[C3] Unwrapped Attention warmup with 15 min cooldown, skip connection no-warmup ...} \\
DAPO & C4$\times$5 / C5$\times$3 & miss & \texttt{[C4] Attention window warmup, no attention skip connection for long sequences ...} \\
SFT & C3$\times$1 / C4$\times$5 / C5$\times$2 & miss & \texttt{[C4] Batched DGL attention} \\
\bottomrule
\end{tabularx}
\end{table}

The interpretation is straightforward. These baselines have learned that staying near attention is the easiest way to keep reward high, even when the genuinely useful direction lies in the optimizer family. SPOD interrupts this bias. Its advantage is not that it forces an artificial uniform distribution, but that it keeps the optimizer family from disappearing under RL pressure.

\subsection{Case 3: Betting on the Head Family Can Artificially Inflate Hit-Based Scores}

The third example shows why hit-based metrics alone can overestimate collapsed policies. This prompt comes from track 1 under a late-stage slice ($1.3$ minutes, $n\_history=8$). The ground-truth accepted change contains both C1 and C4: an algebraic rewrite of XSA that preserves the same computation while improving execution speed.

Because the ground truth itself contains the head family C4, methods that collapse entirely to C4 can still score perfectly on hit-based metrics. In this example, GRPO, STEER, and DAPO all generate 8/8 C4 candidates and all achieve Hit@1; SFT and SPOD also hit (Table~\ref{tab:case3}). The key point is not which method wins this example, but why the collapsed baselines can look strong at all: if the head family is common enough, repeatedly emitting it can directly raise Hit@1 and Hit@8, even when the candidate set has already degenerated semantically.

\begin{table}[htbp]
\centering
\normalsize
\caption{Case 3: a collapsed policy can score well simply by betting on the head family.}
\label{tab:case3}
\begin{tabular*}{\textwidth}{@{\extracolsep{\fill}}llc}
\toprule
Arm & Family distribution over 8 candidates & Hit@1 \\
\midrule
SPOD ($\beta{=}1.0$) & C1$\times$4 / C3$\times$1 / C4$\times$3 & \checkmark \\
GRPO & C4$\times$8 & \checkmark \\
STEER & C4$\times$8 & \checkmark \\
DAPO & C4$\times$8 & \checkmark \\
SFT & C1$\times$1 / C3$\times$1 / C4$\times$5 / C5$\times$1 & \checkmark \\
\bottomrule
\end{tabular*}
\end{table}

This is exactly the kind of behavior that can mask collapse if one only reports hit-based scores. The baselines are not demonstrating healthier search; they are benefiting from a strategy that keeps emitting the head family because that family is common and reward-friendly. This case is therefore direct evidence for why the main paper reports distributional alignment, coverage, and method-diversity metrics in addition to Hit@k.

\subsection{Case 4: Even a Basic High-Gain Optimizer Change Is Missed by the Collapsed Baselines}

The fourth example shows how damaging this bias can become in practice. This prompt comes from track 2 under the cross-track slice at an almost initial state ($29.3$ minutes, $n\_history=1$). The ground-truth accepted direction is a basic optimizer change in C2: adding standard weight decay, with gain $4.096$ minutes, one of the largest gains in the batch and labeled strong-tier.

Such a basic, high-gain optimizer direction should be easy to recover. Yet the RL baselines systematically miss it. GRPO, STEER, and DAPO each allocate five of eight candidates to C4 and generate no C2 candidate at all. SFT still preserves one C2 candidate but does not rank it first. Only SPOD both retains C2 and hits the ground truth at top-1 (Table~\ref{tab:case4}).

\begin{table}[htbp]
\centering
\normalsize
\caption{Case 4: even a basic high-gain optimizer change is systematically missed by the collapsed baselines.}
\label{tab:case4}
\setlength{\tabcolsep}{3pt}
\begin{tabularx}{\textwidth}{@{}>{\raggedright\arraybackslash}p{0.17\textwidth}>{\raggedright\arraybackslash}p{0.24\textwidth}c>{\raggedright\arraybackslash}X@{}}
\toprule
Arm & Family distribution & Hit@1 & First line of top-scored candidate \\
\midrule
SPOD ($\beta{=}1.0$) & C1$\times$2 / C2$\times$1 / C3$\times$3 / C4$\times$2 & \checkmark & \texttt{[C2] Single step SGD, increased step size and mini-batch size, decreased warmup steps ...} \\
GRPO & C1$\times$1 / C3$\times$2 / C4$\times$5 & miss & \texttt{[C4] Faster update attention weights, use long sequence embeddings, shorter training steps ...} \\
STEER & C1$\times$1 / C3$\times$2 / C4$\times$5 & miss & \texttt{[C3] Increase learning rate of RoPE layers by 10x and lower number of k-args for RoPE} \\
DAPO & C1$\times$1 / C3$\times$2 / C4$\times$5 & miss & \texttt{[C4] Single step attention} \\
SFT & C1$\times$1 / C2$\times$1 / C3$\times$2 / C4$\times$4 & miss & \texttt{[C3] Merged batched attention with gradient attention} \\
\bottomrule
\end{tabularx}
\end{table}

This example is especially revealing because the prompt contains only one history item. The baselines therefore cannot plausibly be explained away as following a long, misleading attention-heavy context. Instead, they exhibit a learned default bias: once RL has reshaped the policy, generating attention-related candidates becomes the easiest general-purpose strategy, even when the prompt calls for a basic optimizer change. This is precisely why such methods underperform in the multi-round setting. If even an early, high-gain optimizer direction fails to survive candidate generation, repeated iteration will keep circling around the same head family and lose long-horizon reward.

\subsection{Connection to the Multi-Round Results}

Together, these four cases explain the multi-round results in the previous section. In the replay, GRPO-like methods can obtain some early reward from the head family, but once the high-yield directions in that family are consumed, they continue allocating mass back to the same family, effectively spending later rounds on homogeneous repetition. The single-prompt cases show the same mechanism in local form: the baselines do not merely rank slightly worse, but compress the candidate space into C4 so early that non-head yet high-yield families such as C2 often disappear altogether.

SPOD improves long-horizon performance because it changes this behavior at the source. It prevents RL from turning candidate generation into repeated head-family hacking, preserves the presence of useful non-head families in the pool, and therefore makes it possible for those families to keep contributing reward over repeated iteration.

%% file: sections/appendix_three_layer.tex
\section{Three-Layer Training Dynamics of Diversity}
\label{app:three_layer}

This appendix asks a direct question. When reinforcement learning reduces the
diversity of a policy, at which level does the loss actually happen? To answer it, 
we attach a lightweight diversity probe to the reward pipeline and record
diversity throughout training. The probe only reads each generation batch. It
does not touch the reward, the advantage, or the sampling state, so training
itself is left unchanged. We separate diversity into three layers.

\begin{itemize}
\item \textbf{Token layer.} Token-level entropy of the policy output distribution.
\item \textbf{Surface layer.} Lexical variation of the text, measured on the
      clean evaluation set by word-level Distinct-2, the TF-IDF Vendi score, and
      mean pairwise cosine distance (higher is more diverse); the training monitor
      additionally tracks Self-BLEU (lower is more diverse).
\item \textbf{Method layer.} Variation of the idea that a response proposes. Every
      response is mapped to one of five idea categories (Sys/Kernel, Optimizer,
      Architecture, Attention, Loss). On these categories we track the category
      entropy, the top-1 category share, and the number of categories covered.
\end{itemize}

All seven methods use the same five-category scheme, so the method-layer axis is
comparable across methods. Each run is monitored over 50 optimization steps.

\begin{figure}[htbp]
\centering
\includegraphics[width=0.9\textwidth]{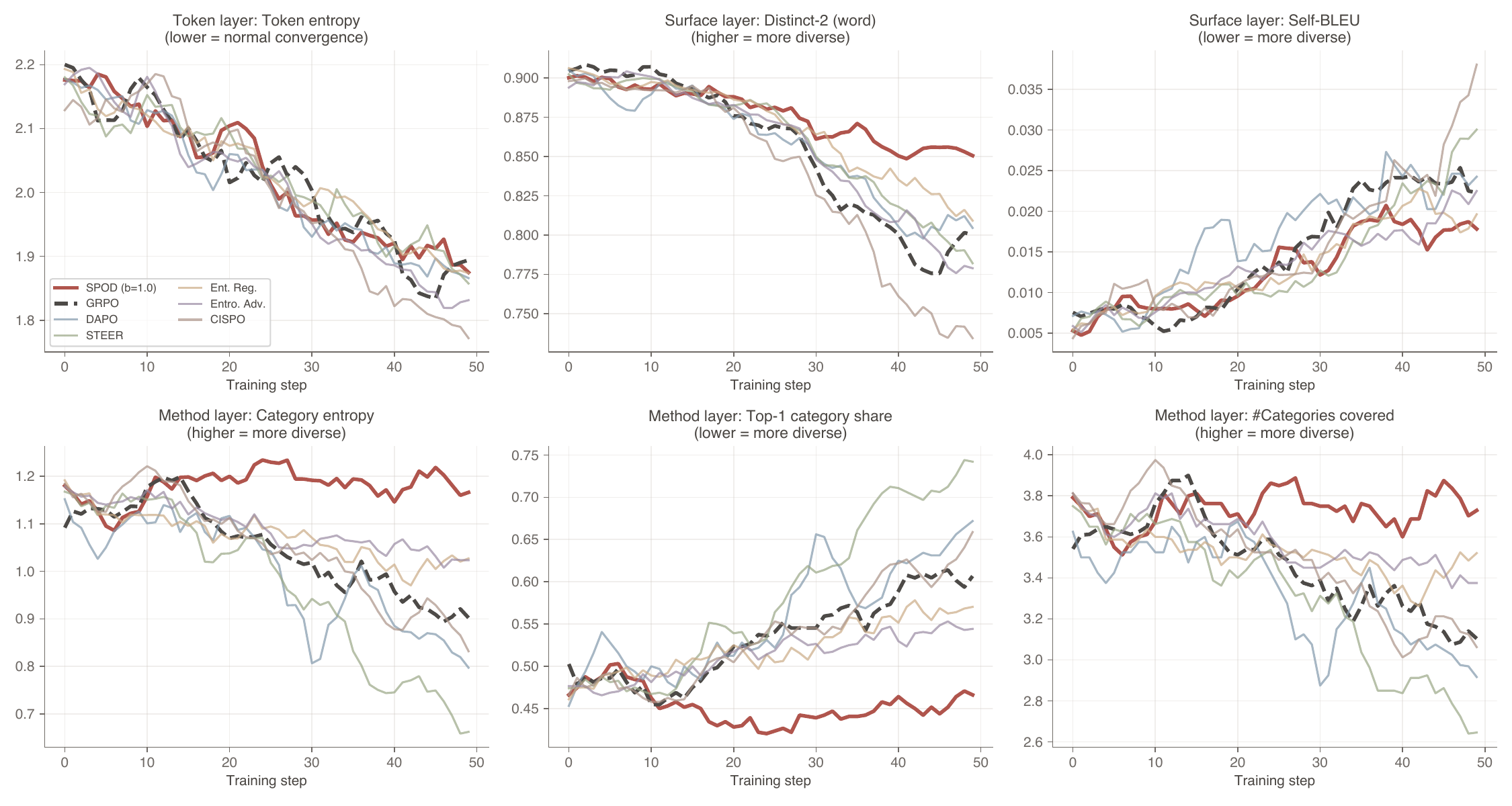}
\caption{Six diversity signals over 50 training steps, grouped by layer. SPOD is
the solid red curve and GRPO is the dashed black curve. Curves use a centered
width-5 moving average. The token layer (top left) and the surface layer (top
middle and top right) show little separation between methods. The method layer
(bottom row) is where the methods split apart, and only SPOD stays diverse.}
\label{fig:three_layer}
\end{figure}

\paragraph{The three layers behave very differently.}
Figure~\ref{fig:three_layer} plots the six signals over training. At the token layer, every method falls by 8 to 22 percent, with GRPO decreasing the least at 8.4 percent. This
is ordinary entropy convergence and shows no diversity-relevant separation between
methods. At the surface layer the methods narrow only mildly. SPOD keeps the most
lexical variation, with Distinct-2 dropping about 2 percent against 3 to 8 percent
for the baselines, and Self-BLEU stays low for all methods. The method layer is the watershed. SPOD
is the only method whose category entropy stays flat and even rises, whose top-1
share stays the lowest, and whose category coverage stays the highest. Every
baseline moves the other way. Category entropy falls, one category swells, and
coverage shrinks.

\begin{figure}[htbp]
\centering
\includegraphics[width=0.9\textwidth]{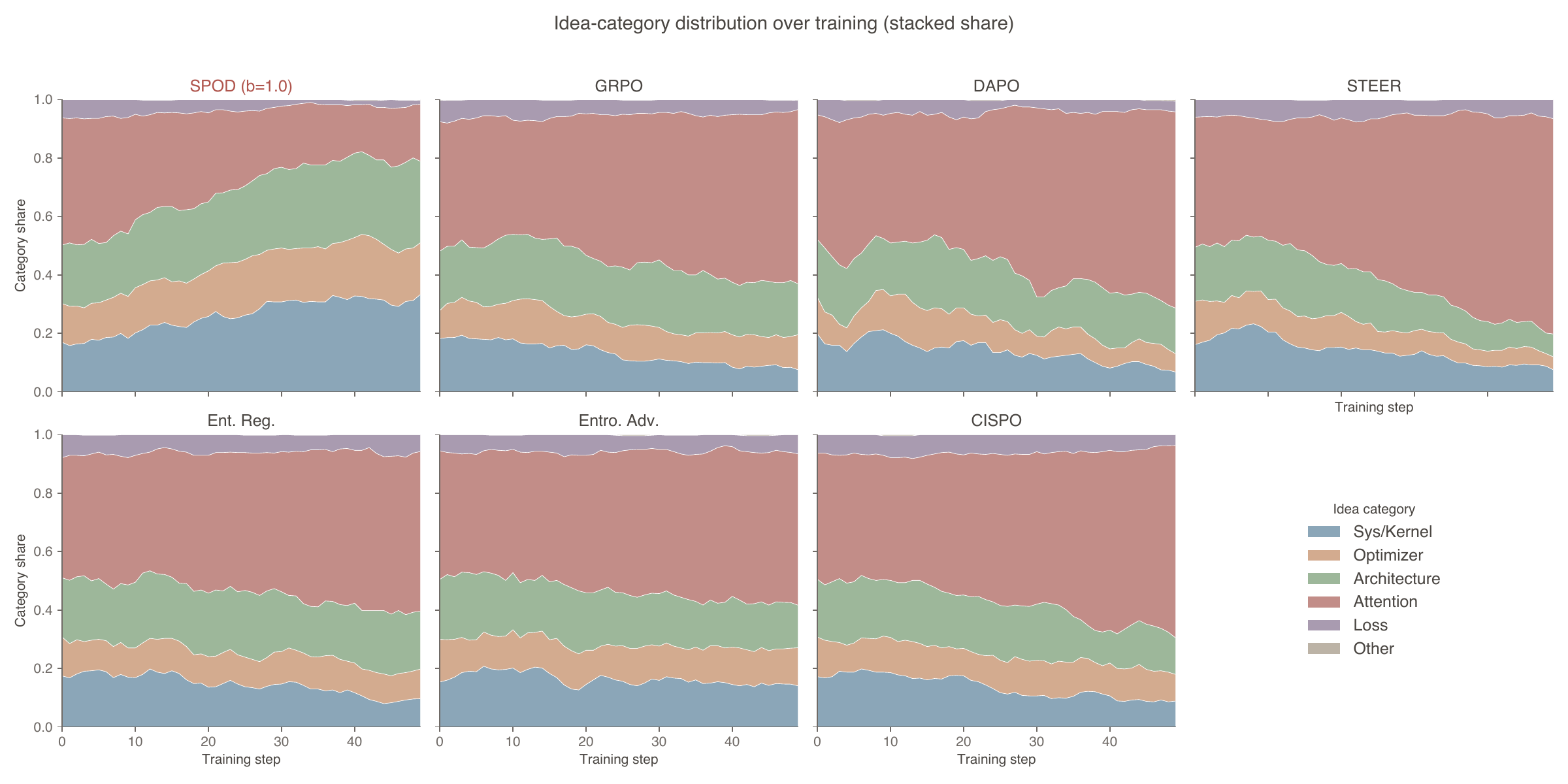}
\caption{Stacked idea-category share over training. A single band that keeps
widening indicates collapse onto one idea. SPOD (top left) is the only method
whose distribution stays balanced and in fact spreads toward the smaller
categories.}
\label{fig:category_share}
\end{figure}

\paragraph{The collapse is a drift toward one idea.}
Figure~\ref{fig:category_share} shows the idea-category mix as a stacked share
over training. For every baseline one band grows steadily wider, most visibly the
Attention category, while the other ideas fade. STEER and CISPO show the sharpest
concentration. SPOD behaves in the opposite way. Its dominant band shrinks and
probability mass flows into the under-used categories, so the mix becomes more
balanced as training proceeds.

\paragraph{Quantitative summary.}
Table~\ref{tab:three_layer_method} reports the method-layer change on the clean
evaluation set relative to the SFT baseline. SPOD is the only method that moves in
the diverse direction on all three signals. Table~\ref{tab:three_layer_tok}
reports the token and surface layers.

\begin{table}[htbp]
\centering
\caption{Method-layer change on the clean evaluation set relative to the SFT
baseline. Arrows give the diverse direction. SPOD is the only method that improves
on all three signals.}
\label{tab:three_layer_method}
\begin{tabular*}{\textwidth}{@{\extracolsep{\fill}}lccc}
\toprule
Method & Category entropy $\uparrow$ & Top-1 share $\downarrow$ & \#Categories $\uparrow$ \\
\midrule
SPOD ($\beta{=}1.0$) & $\mathbf{+6.4\%}$ & $\mathbf{-9.1\%}$ & $\mathbf{+2.9\%}$ \\
GRPO & $-24.3\%$ & $+29.1\%$ & $-18.4\%$ \\
DAPO & $-22.4\%$ & $+27.3\%$ & $-16.5\%$ \\
STEER & $-21.1\%$ & $+26.4\%$ & $-15.5\%$ \\
Ent.\ Reg. & $-26.6\%$ & $+34.5\%$ & $-18.4\%$ \\
Entro.\ Adv. & $-9.4\%$ & $+11.8\%$ & $-6.8\%$ \\
CISPO & $-28.6\%$ & $+33.6\%$ & $-21.4\%$ \\
\bottomrule
\end{tabular*}
\end{table}

\begin{table}[htbp]
\centering
\caption{Token- and surface-layer change. Token entropy is each arm's training-log
endpoint change; the surface metrics are clean-eval changes relative to SFT. SPOD
narrows the least at the surface layer.}
\label{tab:three_layer_tok}
\begin{tabular*}{\textwidth}{@{\extracolsep{\fill}}lcccc}
\toprule
Method & Token entropy & Distinct-2 (word) & Vendi (TF-IDF) & Pairwise \\
\midrule
SPOD ($\beta{=}1.0$) & $-19.1\%$ & $\mathbf{-2.0\%}$ & $\mathbf{-1.8\%}$ & $\mathbf{-3.9\%}$ \\
GRPO & $-8.4\%$ & $-3.5\%$ & $-2.3\%$ & $-4.6\%$ \\
DAPO & $-18.0\%$ & $-4.7\%$ & $-4.8\%$ & $-6.7\%$ \\
STEER & $-17.7\%$ & $-3.6\%$ & $-2.7\%$ & $-4.6\%$ \\
Ent.\ Reg. & $-15.3\%$ & $-5.6\%$ & $-3.8\%$ & $-5.7\%$ \\
Entro.\ Adv. & $-15.3\%$ & $-2.9\%$ & $\mathbf{-1.8\%}$ & $-4.0\%$ \\
CISPO & $-21.9\%$ & $-7.5\%$ & $-5.7\%$ & $-7.7\%$ \\
\bottomrule
\end{tabular*}
\end{table}

\paragraph{Takeaway.}
Diversity collapse under reinforcement learning is not mainly a token or surface
effect: at the token layer all methods converge similarly, and at the surface
layer they narrow only slightly. The loss is concentrated at the method layer,
where a policy keeps writing lexically varied text while quietly narrowing to a
single kind of idea. Among the seven methods studied, only SPOD resists this
method-layer collapse, and the probe shows it doing so during training rather than
only at the final checkpoint.

%% file: sections/appendix_annotation.tex
\section{Robustness to Method-Category Annotation}
\label{app:annot}

SPOD uses method-category labels to correct category-level reward bias, so a natural question is how sensitive it is to the quality of those labels. This question matters especially in the industrial setting, where no ground-truth taxonomy is available and the category signal must be inferred automatically. In this appendix, we test that sensitivity from four angles: weaker category extractors (Section~\ref{app:annot:agree}), noisy labels with full SPOD re-runs (Section~\ref{app:annot:noise}), vocabulary granularity (Section~\ref{app:annot:gran}), and perturbations of the $\phi$ table (Section~\ref{app:annot:phi}). Across all four tests, the same conclusion emerges. SPOD does not require a precise or fragile taxonomy. It only needs a category partition that is coarse but still informative.

\subsection{Q1: Weaker category extractors}
\label{app:annot:agree}

We first test whether SPOD still helps when the category signal is replaced by a much weaker one. This is important because, in practice, method categories may be extracted imperfectly rather than provided as clean labels. To stress-test this dependence, we construct two substantially weaker method-category extractors and use them in place of the category assignment used in the main experiments: (i) $X_{\text{cent}}$, a TF--IDF nearest-centroid classifier, and (ii) $X_{\text{kw}}$, a keyword-scoring rule-based extractor.



We then re-evaluate the same held-out outputs using category assignments produced by each weaker extractor. Because different extractors change the measurement scale, the absolute metric values are not the main point of comparison. The key question is whether SPOD still improves over GRPO in the same direction as in the main results. Table~\ref{tab:annot-sign} shows that it does. For every metric that still provides a meaningful comparison, the direction of the SPOD-over-GRPO improvement is unchanged. The only exception is All-8-same-category, which becomes $0.000$ for both methods under both weak extractors and therefore no longer distinguishes between them.

\begin{table}[!ht]
\centering
\caption{Held-out metrics recomputed after replacing the category field with a weaker extractor. Entries are $\Delta = \text{SPOD} - \text{GRPO}$. ``Sign'' counts how many of the two weaker extractors preserve the direction of the main result.}
\label{tab:annot-sign}
\normalsize
\begin{tabular}{llcccc}
\toprule
Metric & Dir. & Main exp. field & $X_{\text{cent}}$ & $X_{\text{kw}}$ & Sign \\
\midrule
\multicolumn{6}{l}{\emph{Quality}} \\
Hit@1                    & $\uparrow$   & $+0.218$ & $+0.130$ & $+0.130$ & 2/2 \\
Hit@8                    & $\uparrow$   & $+0.217$ & $+0.087$ & $+0.130$ & 2/2 \\
Gold-category coverage   & $\uparrow$   & $+0.305$ & $+0.087$ & $+0.207$ & 2/2 \\
TV to library prior      & $\downarrow$ & $-0.375$ & $-0.139$ & $-0.070$ & 2/2 \\
\midrule
\multicolumn{6}{l}{\emph{Diversity}} \\
distinct ideas         & $\uparrow$   & $+0.786$ & $+0.464$ & $+0.250$ & 2/2 \\
Top-1 category share     & $\downarrow$ & $-0.286$ & $-0.170$ & $-0.118$ & 2/2 \\
Mean max same-category   & $\downarrow$ & $-1.500$ & $-0.679$ & $-0.714$ & 2/2 \\
All-8-same-category      & $\downarrow$ & $-0.071$ & $\ \ 0.000$ & $\ \ 0.000$ & n/a \\
\midrule
\multicolumn{5}{l}{\textbf{Total informative directions preserved}} & \textbf{14/14} \\
\bottomrule
\end{tabular}
\end{table}

These results show that the main GRPO-to-SPOD improvement does not depend on a high-precision category extractor. Even after replacing the category signal with much weaker alternatives, the quality and diversity gains remain in the same direction.




\subsection{Q2: Does SPOD remain useful when the method labels are noisy?}
\label{app:annot:noise}

SPOD relies on method-category labels to apply category-aware reward correction. A natural concern, therefore, is whether this dependence makes it brittle: if the labels are imperfect, does SPOD fail quickly, or does it degrade gradually as the category signal becomes noisier?

To test this, we deliberately corrupt the method labels and then re-run SPOD using the corrupted annotations. Concretely, for each example, we keep the original label with probability $1-\varepsilon$, and with probability $\varepsilon$ we replace it with a different category label. This lets us control how much category information remains available to SPOD.

\paragraph{What should happen under random label noise?}
SPOD's correction is category-dependent. If the labels are corrupted at random, the correction should not disappear all at once; instead, it should be progressively diluted because the category signal becomes less reliable. Under the simple noise model above, the expected correction for category $c$ becomes
$$
\phi_{\mathrm{eff}}(c)
=
\left(1-\frac{K}{K-1}\varepsilon\right)\phi(c),
$$
where $K$ is the number of categories. In other words, random label noise mainly scales down the original correction signal.

The correction vanishes when
$$
\varepsilon^\star = \frac{K-1}{K}.
$$
In our main setting, $K=5$, so $\varepsilon^\star = 0.8$. This gives a clear prediction: as $\varepsilon$ increases, SPOD should gradually move back toward the GRPO baseline. It should not collapse as soon as a modest amount of label noise is introduced.

Table~\ref{tab:annot-noise}(a) shows exactly this pattern. As the labels become noisier, SPOD weakens smoothly rather than failing abruptly. Across $\varepsilon \in [0,0.6]$, it remains better than GRPO on all eight metrics. At $\varepsilon^\star = 0.8$, where the expected correction is zero, the results return to nearly the GRPO level. Overall, these results suggest that noisy labels mainly reduce the strength of SPOD's correction, rather than turning it into a harmful signal.

\begin{table}[!ht]
\centering
\caption{Robustness of SPOD to noisy method labels and to category granularity, compared with a common GRPO reference row. Panel (a) corrupts the method labels and re-runs SPOD. Its \textit{Scale} column reports how much of the original correction signal is expected to remain under that noise level. Panel (b) changes the number of method categories $K$. There, \textit{Scale} reports the relative strength of the resulting correction signal, normalized so that the main-text setting ($K=5$) equals $100\%$.}
\label{tab:annot-noise}
\normalsize
\setlength{\tabcolsep}{2.5pt}
\begin{tabular}{lc@{\hskip .8em}cccc@{\hskip 1.2em}cccc}
\toprule
& & \multicolumn{4}{c}{Quality} & \multicolumn{4}{c}{Diversity} \\
\cmidrule(lr){3-6}\cmidrule(lr){7-10}
Setting & Scale & Hit@1$\uparrow$ & Hit@8$\uparrow$ & Cov.$\uparrow$
& TV$_{\text{lib}}\downarrow$ & \#Ideas$\uparrow$ & Top-1$\downarrow$
& MaxSame$\downarrow$ & AllSame$\downarrow$ \\
\midrule
GRPO & --- & 0.217 & 0.696 & 0.565 & 0.498 & 3.000 & 0.621 & 5.071 & 0.071 \\
\midrule
\multicolumn{10}{l}{\emph{(a) Noisy method categories; scale $=\phi_{\text{eff}}$}} \\
$\varepsilon=0.0$ & 1.000 & \textbf{0.435} & \textbf{0.913} & \textbf{0.870} & \textbf{0.123} & \textbf{3.786} & \textbf{0.335} & \textbf{3.571} & \textbf{0.000} \\
$\varepsilon=0.1$ & 0.875 & 0.435 & 0.913 & 0.826 & 0.146 & 3.679 & 0.375 & 3.607 & 0.000 \\
$\varepsilon=0.2$ & 0.750 & 0.435 & 0.870 & 0.783 & 0.155 & 3.714 & 0.353 & 3.500 & 0.000 \\
$\varepsilon=0.3$ & 0.625 & 0.435 & 0.913 & 0.870 & 0.213 & 3.571 & 0.357 & 3.857 & 0.000 \\
$\varepsilon=0.4$ & 0.500 & 0.391 & 0.870 & 0.783 & 0.235 & 3.643 & 0.362 & 3.607 & 0.000 \\
$\varepsilon=0.5$ & 0.375 & 0.304 & 0.826 & 0.783 & 0.347 & 3.393 & 0.460 & 4.143 & 0.036 \\
$\varepsilon=0.6$ & 0.250 & 0.261 & 0.783 & 0.652 & 0.404 & 3.143 & 0.518 & 4.714 & 0.036 \\
$\varepsilon^\star=0.8$ & 0.000 & 0.261 & 0.696 & 0.565 & 0.489 & 3.036 & 0.607 & 5.036 & 0.071 \\
$\varepsilon=1.0$ & -0.250 & 0.130 & 0.522 & 0.435 & 0.718 & 2.393 & 0.862 & 6.143 & 0.143 \\
\midrule
\multicolumn{10}{l}{\emph{(b) Vocabulary granularity; row label $=K$}} \\
1   & \ \ 0\% & 0.261 & 0.696 & 0.565 & 0.489 & 3.000 & 0.616 & 5.000 & 0.071 \\
2   & \ 97\% & 0.435 & 0.870 & 0.870 & 0.143 & 3.750 & 0.384 & 3.786 & 0.000 \\
3   & 108\% & 0.391 & 0.913 & 0.870 & 0.159 & 3.679 & 0.375 & 3.679 & 0.000 \\
5   & 100\% & \textbf{0.435} & \textbf{0.913} & \textbf{0.870} & \textbf{0.123} & \textbf{3.786} & \textbf{0.335} & \textbf{3.571} & \textbf{0.000} \\
8   & \ 88\% & 0.391 & \textbf{0.913} & 0.826 & 0.203 & 3.571 & 0.384 & 3.714 & 0.000 \\
20  & \ 41\% & 0.391 & 0.826 & 0.696 & 0.363 & 3.357 & 0.451 & 4.286 & 0.036 \\
40  & \ 21\% & 0.261 & 0.739 & 0.609 & 0.454 & 3.107 & 0.558 & 4.714 & 0.036 \\
184 & \ \ 5\% & 0.217 & 0.696 & 0.609 & 0.492 & 2.964 & 0.629 & 5.107 & 0.071 \\
\bottomrule
\end{tabular}
\end{table}

\subsection{Q3: Vocabulary granularity}
\label{app:annot:gran}

We next test whether SPOD only works at one carefully chosen category granularity. This matters because any practical taxonomy requires a design choice about how coarse or fine the category space should be. If SPOD depended on a single special vocabulary size, its usefulness would be much more limited.

The main text uses a five-category vocabulary. Table~\ref{tab:annot-noise}(b) shows that changing the number of categories mainly weakens SPOD toward GRPO rather than below it. At the coarsest extreme, $K=1$, no category structure remains and SPOD moves back to the GRPO regime. At the finest extreme considered here, $K=184$, the category space becomes so fine that the correction signal is spread too thinly, and the results again move back toward GRPO. The five-category setting used in the main text therefore does not behave like a fragile special case. It lies inside a fairly broad effective range, and nearby coarsening or refinement leaves the qualitative conclusion unchanged.

\subsection{Q4: Sensitivity to the \texorpdfstring{$\phi$}{phi} table}
\label{app:annot:phi}

Finally, we test whether SPOD depends on a finely calibrated correction table $\phi$. Here, the $\phi$ table is the lookup table that assigns each category its correction score in the reward debiasing step. Even if the category assignments themselves are reasonable, the method could still be fragile if it required very precise numerical values for these scores.

Table~\ref{tab:annot-phi} compares several informative variants of $\phi$, including two independently calibrated tables, a sign-only version, and a sparse one-category version, against two controls: a constant $\phi$ and a random permutation. The goal is to see whether SPOD needs one very specific calibration, or whether it works as long as the table preserves the broad structure of category-level bias.

\begin{table}[!ht]
\centering
\caption{Results under different $\phi$ tables. The first four rows are informative variants, and the last two are controls. Random permutation is averaged over 30 draws.}
\label{tab:annot-phi}
\normalsize
\setlength{\tabcolsep}{1.5pt}
\begin{tabular}{l@{\hskip .8em}cccc@{\hskip 1.2em}cccc}
\toprule
& \multicolumn{4}{c}{Quality} & \multicolumn{4}{c}{Diversity} \\
\cmidrule(lr){2-5}\cmidrule(lr){6-9}
$\phi$ table & Hit@1$\uparrow$ & Hit@8$\uparrow$ & Cov.$\uparrow$
& TV$_{\text{lib}}\downarrow$ & \#Ideas$\uparrow$ & Top-1$\downarrow$
& MaxSame$\downarrow$ & AllSame$\downarrow$ \\
\midrule
GRPO (no correction) & 0.217 & 0.696 & 0.565 & 0.498 & 3.000 & 0.621 & 5.071 & 0.071 \\
\midrule
Calibrated (main text)                & \textbf{0.435} & \textbf{0.913} & \textbf{0.870} & \textbf{0.123} & \textbf{3.786} & \textbf{0.335} & \textbf{3.571} & \textbf{0.000} \\
Earlier calibration                   & 0.391 & 0.870 & 0.783 & 0.238 & 3.643 & 0.433 & 3.964 & 0.000 \\
$\mathrm{sign}(\phi)$ only            & 0.435 & 0.913 & 0.783 & 0.241 & 3.714 & 0.424 & 3.821 & 0.000 \\
\texttt{C4} only ($\phi=1$, else $0$) & 0.261 & 0.826 & 0.739 & 0.396 & 3.286 & 0.496 & 4.536 & 0.036 \\
\midrule
\emph{Control:} $\phi \equiv 1$       & 0.217 & 0.826 & 0.696 & 0.431 & 3.357 & 0.558 & 4.607 & 0.036 \\
\emph{Control:} random perm.          & 0.217 & 0.696 & 0.609 & 0.512 & 3.036 & 0.649 & 4.964 & 0.071 \\
\bottomrule
\end{tabular}
\end{table}

Table~\ref{tab:annot-phi} leads to a clear conclusion. SPOD does not depend on a highly specific calibration of the $\phi$ table. As long as the table preserves meaningful category-level structure, performance stays within a similar range. Once that structure is removed, as in the constant and permuted controls, the gains largely disappear. This shows that SPOD relies on the coarse pattern of category-level bias rather than on finely tuned numerical values.

Across Tables~\ref{tab:annot-sign}--\ref{tab:annot-phi}, the overall conclusion is consistent. SPOD does not require a precise or brittle taxonomy. Its gains persist under weaker category extractors, noisy method labels, changes in vocabulary granularity, and substantial perturbations of the $\phi$ table. As the category signal becomes less informative, SPOD degrades smoothly toward GRPO rather than failing abruptly or dropping below it.

%% file: sections/appendix_multiseed.tex
\section{Multi-Seed Results and Statistical Tests}
\label{app:multiseed}

We repeat the overall public-benchmark evaluation for all seven post-training
methods across five random seeds $\{42, 211, 985, 43, 44\}$. Every run uses the
same SFT initialization, reward model, training prompts, and 28-task test set,
and differs only in the RL objective or the diversity-preserving mechanism. Each
task has eight generated candidates. Hit@1, Hit@8, and Coverage use the 23 tasks
that carry a ground-truth method label, while the four diversity metrics use all
28 tasks. TV measures the deviation from the reference library category
distribution. Arrows in the headers give the preferred direction, and bold marks
the best method for each metric. Metric abbreviations follow
Table~\ref{tab:rustc_diversity}.

The five per-seed tables below list the raw scores. On every seed SPOD attains the
best value on seven of the eight metrics, including all four diversity metrics
together with TV, Coverage, and Hit@8, and on three of the five seeds it is best on
all eight. 

\begin{center}
\textbf{Seed 42}\\[2pt]
\normalsize
\setlength{\tabcolsep}{5pt}
\begin{tabular*}{\textwidth}{@{\extracolsep{\fill}}lcccccccc}
\toprule
Method & Hit@1$\uparrow$ & Hit@8$\uparrow$ & Cov.$\uparrow$ & TV$\downarrow$ & Dist.$\uparrow$ & TopFam$\downarrow$ & MaxInFam$\downarrow$ & Coll.$\downarrow$ \\
\midrule
SPOD ($\beta{=}1.0$) & \textbf{0.435} & \textbf{0.913} & \textbf{0.870} & \textbf{0.123} & \textbf{3.786} & \textbf{0.335} & \textbf{3.571} & \textbf{0.000} \\
GRPO & 0.217 & 0.696 & 0.565 & 0.498 & 3.000 & 0.621 & 5.071 & 0.071 \\
DAPO & 0.304 & 0.739 & 0.641 & 0.498 & 3.071 & 0.612 & 5.000 & 0.071 \\
STEER & 0.217 & 0.739 & 0.641 & 0.475 & 3.107 & 0.603 & 4.964 & 0.071 \\
Ent.\ Reg. & 0.217 & 0.783 & 0.652 & 0.524 & 3.000 & 0.643 & 5.286 & 0.071 \\
Entro.\ Adv. & 0.261 & 0.826 & 0.707 & 0.395 & 3.429 & 0.522 & 4.393 & 0.036 \\
CISPO & 0.217 & 0.696 & 0.565 & 0.515 & 2.893 & 0.643 & 5.250 & 0.071 \\
\bottomrule
\end{tabular*}
\end{center}

\begin{center}
\textbf{Seed 211}\\[2pt]
\normalsize
\setlength{\tabcolsep}{5pt}
\begin{tabular*}{\textwidth}{@{\extracolsep{\fill}}lcccccccc}
\toprule
Method & Hit@1$\uparrow$ & Hit@8$\uparrow$ & Cov.$\uparrow$ & TV$\downarrow$ & Dist.$\uparrow$ & TopFam$\downarrow$ & MaxInFam$\downarrow$ & Coll.$\downarrow$ \\
\midrule
SPOD ($\beta{=}1.0$) & \textbf{0.478} & \textbf{0.783} & \textbf{0.728} & \textbf{0.181} & \textbf{3.679} & \textbf{0.335} & \textbf{3.571} & \textbf{0.000} \\
GRPO & 0.261 & 0.696 & 0.565 & 0.484 & 3.036 & 0.612 & 5.000 & 0.071 \\
DAPO & 0.261 & 0.739 & 0.609 & 0.440 & 3.107 & 0.567 & 4.786 & 0.036 \\
STEER & 0.391 & 0.783 & 0.674 & 0.350 & 3.321 & 0.478 & 4.107 & 0.000 \\
Ent.\ Reg. & 0.304 & 0.652 & 0.522 & 0.569 & 2.750 & 0.692 & 5.607 & 0.071 \\
Entro.\ Adv. & 0.391 & 0.783 & 0.663 & 0.417 & 3.321 & 0.545 & 4.571 & 0.036 \\
CISPO & 0.261 & 0.696 & 0.565 & 0.542 & 2.929 & 0.661 & 5.393 & 0.071 \\
\bottomrule
\end{tabular*}
\end{center}

\begin{center}
\textbf{Seed 985}\\[2pt]
\normalsize
\setlength{\tabcolsep}{5pt}
\begin{tabular*}{\textwidth}{@{\extracolsep{\fill}}lcccccccc}
\toprule
Method & Hit@1$\uparrow$ & Hit@8$\uparrow$ & Cov.$\uparrow$ & TV$\downarrow$ & Dist.$\uparrow$ & TopFam$\downarrow$ & MaxInFam$\downarrow$ & Coll.$\downarrow$ \\
\midrule
SPOD ($\beta{=}1.0$) & 0.348 & \textbf{0.870} & \textbf{0.826} & \textbf{0.114} & \textbf{3.750} & \textbf{0.357} & \textbf{3.679} & \textbf{0.000} \\
GRPO & \textbf{0.435} & 0.739 & 0.609 & 0.480 & 3.036 & 0.598 & 4.929 & 0.071 \\
DAPO & 0.304 & 0.696 & 0.587 & 0.484 & 3.036 & 0.612 & 5.036 & 0.071 \\
STEER & 0.304 & 0.739 & 0.630 & 0.538 & 3.000 & 0.656 & 5.357 & 0.071 \\
Ent.\ Reg. & 0.174 & 0.739 & 0.620 & 0.524 & 3.107 & 0.638 & 5.214 & 0.071 \\
Entro.\ Adv. & 0.348 & 0.826 & 0.707 & 0.390 & 3.393 & 0.518 & 4.429 & 0.000 \\
CISPO & 0.217 & 0.783 & 0.663 & 0.480 & 3.071 & 0.607 & 4.964 & 0.071 \\
\bottomrule
\end{tabular*}
\end{center}

\begin{center}
\textbf{Seed 43}\\[2pt]
\normalsize
\setlength{\tabcolsep}{5pt}
\begin{tabular*}{\textwidth}{@{\extracolsep{\fill}}lcccccccc}
\toprule
Method & Hit@1$\uparrow$ & Hit@8$\uparrow$ & Cov.$\uparrow$ & TV$\downarrow$ & Dist.$\uparrow$ & TopFam$\downarrow$ & MaxInFam$\downarrow$ & Coll.$\downarrow$ \\
\midrule
SPOD ($\beta{=}1.0$) & 0.304 & \textbf{0.870} & \textbf{0.815} & \textbf{0.118} & \textbf{3.786} & \textbf{0.371} & \textbf{3.750} & \textbf{0.000} \\
GRPO & \textbf{0.348} & 0.739 & 0.609 & 0.444 & 3.143 & 0.571 & 4.750 & 0.036 \\
DAPO & 0.217 & 0.783 & 0.652 & 0.471 & 3.179 & 0.589 & 4.893 & 0.071 \\
STEER & 0.261 & 0.696 & 0.565 & 0.524 & 2.964 & 0.634 & 5.179 & 0.071 \\
Ent.\ Reg. & 0.304 & 0.783 & 0.685 & 0.475 & 3.107 & 0.603 & 4.964 & 0.071 \\
Entro.\ Adv. & 0.348 & 0.739 & 0.609 & 0.453 & 3.107 & 0.580 & 4.821 & 0.071 \\
CISPO & 0.304 & 0.739 & 0.630 & 0.515 & 3.000 & 0.643 & 5.250 & 0.071 \\
\bottomrule
\end{tabular*}
\end{center}

\begin{center}
\textbf{Seed 44}\\[2pt]
\normalsize
\setlength{\tabcolsep}{5pt}
\begin{tabular*}{\textwidth}{@{\extracolsep{\fill}}lcccccccc}
\toprule
Method & Hit@1$\uparrow$ & Hit@8$\uparrow$ & Cov.$\uparrow$ & TV$\downarrow$ & Dist.$\uparrow$ & TopFam$\downarrow$ & MaxInFam$\downarrow$ & Coll.$\downarrow$ \\
\midrule
SPOD ($\beta{=}1.0$) & \textbf{0.435} & \textbf{0.913} & \textbf{0.870} & \textbf{0.109} & \textbf{3.786} & \textbf{0.379} & \textbf{3.750} & \textbf{0.000} \\
GRPO & 0.217 & 0.739 & 0.609 & 0.511 & 3.036 & 0.638 & 5.214 & 0.071 \\
DAPO & 0.348 & 0.696 & 0.587 & 0.506 & 3.000 & 0.634 & 5.179 & 0.071 \\
STEER & 0.261 & 0.696 & 0.554 & 0.574 & 2.821 & 0.688 & 5.500 & 0.071 \\
Ent.\ Reg. & 0.217 & 0.739 & 0.630 & 0.471 & 3.179 & 0.580 & 4.857 & 0.036 \\
Entro.\ Adv. & 0.261 & 0.826 & 0.696 & 0.413 & 3.321 & 0.540 & 4.536 & 0.036 \\
CISPO & 0.261 & 0.783 & 0.663 & 0.507 & 3.179 & 0.621 & 5.107 & 0.071 \\
\bottomrule
\end{tabular*}
\end{center}

Table~\ref{tab:seed_quality} and Table~\ref{tab:seed_diversity} average these
runs and report the sample standard deviation ($\mathrm{ddof}=1$). SPOD ranks
first by the mean on all eight metrics, and Entro.\ Adv.\ is the runner-up on all
eight. The gap between the best and the second-best mean is large for the
alignment and diversity metrics, for example $+0.146$ on Coverage, $-0.285$ on
TV, and $-0.886$ on Max Within-Family Count.

\begin{table}[htbp]
\centering
\caption{Public-benchmark quality and alignment: mean $\pm$ sample standard deviation over five seeds. Bold marks the best method.}
\label{tab:seed_quality}
\begin{tabular*}{\textwidth}{@{\extracolsep{\fill}}lcccc}
\toprule
Method & Hit@1$\uparrow$ & Hit@8$\uparrow$ & Cov.$\uparrow$ & TV$\downarrow$ \\
\midrule
SPOD ($\beta{=}1.0$) & $\mathbf{0.400}\pm0.071$ & $\mathbf{0.870}\pm0.053$ & $\mathbf{0.822}\pm0.058$ & $\mathbf{0.129}\pm0.029$ \\
GRPO & $0.296\pm0.094$ & $0.722\pm0.024$ & $0.591\pm0.024$ & $0.483\pm0.025$ \\
DAPO & $0.287\pm0.050$ & $0.730\pm0.036$ & $0.615\pm0.030$ & $0.480\pm0.026$ \\
STEER & $0.287\pm0.066$ & $0.730\pm0.036$ & $0.613\pm0.051$ & $0.492\pm0.087$ \\
Ent.\ Reg. & $0.243\pm0.058$ & $0.739\pm0.053$ & $0.622\pm0.061$ & $0.513\pm0.041$ \\
Entro.\ Adv. & $0.322\pm0.058$ & $0.800\pm0.039$ & $0.676\pm0.042$ & $0.414\pm0.025$ \\
CISPO & $0.252\pm0.036$ & $0.739\pm0.043$ & $0.617\pm0.049$ & $0.512\pm0.022$ \\
\bottomrule
\end{tabular*}
\end{table}

\begin{table}[htbp]
\centering
\caption{Public-benchmark diversity: mean $\pm$ sample standard deviation over five seeds. Metric abbreviations follow Table~\ref{tab:rustc_diversity}. Bold marks the best method.}
\label{tab:seed_diversity}
\begin{tabular*}{\textwidth}{@{\extracolsep{\fill}}lcccc}
\toprule
Method & Dist.$\uparrow$ & TopFam$\downarrow$ & MaxInFam$\downarrow$ & Coll.$\downarrow$ \\
\midrule
SPOD ($\beta{=}1.0$) & $\mathbf{3.757}\pm0.047$ & $\mathbf{0.355}\pm0.020$ & $\mathbf{3.664}\pm0.090$ & $\mathbf{0.000}\pm0.000$ \\
GRPO & $3.050\pm0.054$ & $0.608\pm0.025$ & $4.993\pm0.172$ & $0.064\pm0.016$ \\
DAPO & $3.079\pm0.069$ & $0.603\pm0.025$ & $4.979\pm0.149$ & $0.064\pm0.016$ \\
STEER & $3.043\pm0.186$ & $0.612\pm0.081$ & $5.021\pm0.549$ & $0.057\pm0.032$ \\
Ent.\ Reg. & $3.029\pm0.168$ & $0.631\pm0.043$ & $5.186\pm0.294$ & $0.064\pm0.016$ \\
Entro.\ Adv. & $3.314\pm0.125$ & $0.541\pm0.025$ & $4.550\pm0.169$ & $0.036\pm0.025$ \\
CISPO & $3.014\pm0.115$ & $0.635\pm0.021$ & $5.193\pm0.163$ & $0.071\pm0.000$ \\
\bottomrule
\end{tabular*}
\end{table}

We next test whether the advantage of SPOD over the runner-up is statistically
reliable. The comparison uses two-sided paired permutation tests between SPOD and
Entro.\ Adv. The paired unit is a (task, seed) pair, which gives $n=140$ over the
five seeds. For each metric the statistic is the observed difference between the
two methods, oriented so that a positive value favors SPOD. Each of $B=20{,}000$
permutations swaps the two methods within a matched unit and recomputes the
statistic, with
\begin{equation*}
p=\frac{1+\sum_{b=1}^{B}\mathbf{1}\{|\Delta_b^*|\geq|\Delta_{\mathrm{obs}}|\}}{B+1}.
\end{equation*}
TV and Top-Family Share are recomputed from pooled category counts rather than
averaged over per-task differences. The test relies on exchangeability under the
null and does not by itself model a population of independent training seeds. Holm
and Benjamini--Hochberg (BH) corrections are applied across the eight metrics.

\begin{table}[htbp]
\centering
\caption{SPOD versus Entro.\ Adv.\ on the public benchmark: two-sided paired permutation tests with multiplicity-adjusted values. Positive $\Delta$ favors SPOD. A star marks adjusted $p<0.05$.}
\label{tab:seed_tests}
\begin{tabular*}{\textwidth}{@{\extracolsep{\fill}}lccccc}
\toprule
Metric & $\Delta$ & $z$ & $p$ & Holm & BH \\
\midrule
Hit@1 & $+0.0783$ & $+1.53$ & $.1750$ & $.1750$ & $.1750$ \\
Hit@8 & $+0.0696$ & $+2.54$ & $.0194$ & $.0582$ & $.0259^{*}$ \\
Cov. & $+0.1457$ & $+4.48$ & $<.0001$ & $.0004^{*}$ & $<.0001^{*}$ \\
TV & $+0.2846$ & $+9.55$ & $<.0001$ & $.0004^{*}$ & $<.0001^{*}$ \\
Dist. & $+0.4429$ & $+5.33$ & $<.0001$ & $.0004^{*}$ & $<.0001^{*}$ \\
TopFam & $+0.1857$ & $+5.96$ & $<.0001$ & $.0004^{*}$ & $<.0001^{*}$ \\
MaxInFam & $+0.8857$ & $+6.35$ & $<.0001$ & $.0004^{*}$ & $<.0001^{*}$ \\
Coll. & $+0.0357$ & $+2.22$ & $.0640$ & $.1280$ & $.0731$ \\
\bottomrule
\end{tabular*}
\end{table}

Table~\ref{tab:seed_tests} collects the results. Coverage, TV, Distinct Method
Count, Top-Family Share, and Max Within-Family Count are all significant after Holm
and BH correction, with raw permutation $p$ below $0.001$. Hit@8 also clears the
raw and BH criteria. The observed difference is positive for SPOD on every metric.
The strongest corrected evidence therefore concerns distributional alignment and
reduced category concentration, which are the properties SPOD is designed to
improve.

%% file: sections/appendix_limitations.tex
\section{Limitations}
\label{app:limitations}

SPOD relies on method-category labels for category-aware reward correction. On benchmarks where such categories are available, they can be used directly; in new domains without an existing taxonomy, they must be obtained through semantic annotation. Appendix~\ref{app:annot} shows that SPOD degrades gracefully rather than collapsing when the category signal is weaker, labels are noisy, or vocabulary size changes. Still, constructing and validating a useful taxonomy for a new repository domain requires some upfront effort. This is mainly an engineering cost rather than a fundamental limitation, since the annotation pipeline can be built largely through automatic LLM extraction with only light human checking.
